\documentclass{article}

\PassOptionsToPackage{numbers,  sort&compress}{natbib}
\usepackage[final]{neurips_2020}
\usepackage{natbib}

\usepackage{color}
\definecolor{mycitecolor}{rgb}{0,0.08,0.45} 

\usepackage[utf8]{inputenc} 
\usepackage[T1]{fontenc}    
\usepackage[colorlinks=true, citecolor=mycitecolor]{hyperref}
\usepackage{url}            
\usepackage{booktabs}       
\usepackage{amsfonts}       
\usepackage{nicefrac}       
\usepackage{microtype}      
\usepackage{multirow}   
\usepackage{algorithm}
\usepackage{algorithmic}
\usepackage{wrapfig}
\usepackage{graphicx}
\usepackage{comment}
\usepackage{amsmath,amssymb} 
\usepackage{symbols}
\usepackage{multirow}
\usepackage{ctable} 
\usepackage{pifont}
\usepackage{amsmath,amsfonts,bm}

\def\1{\bm{1}}

\def\sp{space}

\DeclareMathAlphabet{\mathsfit}{\encodingdefault}{\sfdefault}{m}{sl}
\SetMathAlphabet{\mathsfit}{bold}{\encodingdefault}{\sfdefault}{bx}{n}

\newcommand*{\ShowNotes}{} 
\definecolor{darkred}{rgb}{0.7,0.1,0.1}
\definecolor{darkgreen}{rgb}{0.1,0.7,0.1}
\definecolor{cyan}{rgb}{0.7,0.0,0.7}
\definecolor{dblue}{rgb}{0.2,0.2,0.8}
\definecolor{maroon}{rgb}{0.76,.13,.28}
\definecolor{burntorange}{rgb}{0.81,.33,0}
\definecolor{tealblue}{rgb}{0.212,0.459, 0.533}

\definecolor{mypink}{rgb}{0.93359375, 0.62109375, 0.83984375}

\definecolor{pp}{rgb}{0.43921569, 0.18823529, 0.62745098}
\definecolor{rr}{rgb}{0.5254902 , 0.00784314, 0.12941176}
\definecolor{bb}{rgb}{0.09019608, 0.23529412, 0.37647059}
\definecolor{yy}{rgb}{0.49803922, 0.3372549 , 0.0}
\definecolor{gg}{rgb}{0.02352941, 0.3372549 , 0.17647059}

\ifdefined\ShowNotes
  \newcommand{\colornote}[3]{{\color{#1}\bf{#2: #3}\normalfont}}
\else
  \newcommand{\colornote}[3]{}
\fi

\newcommand{\eat}[1]{} 

\usepackage[multiple]{footmisc}

\definecolor{mybrown}{rgb}{0.87058824, 0.56078431, 0.01960784}
\definecolor{myblue}{rgb}{0.3372549 , 0.70588235, 0.91372549}
\definecolor{mypurple}{rgb}{0.8, 0.47058824, 0.7372549 }
\definecolor{myorange}{rgb}{0.835, 0.368, 0}
\definecolor{mygreen}{rgb}{0.00784314, 0.61960784, 0.45098039}
\definecolor{mygt}{rgb}{0.0078125 , 0.57421875, 0.40625}
\definecolor{mysp}{rgb}{0.84765625, 0.515625  , 0.0234375}

\title{
Not All Unlabeled Data are Equal: \\Learning to Weight Data in Semi-supervised Learning
}

\author{%
  Zhongzheng Ren\thanks{Indicates equal contribution}, \;
  \addtocounter{footnote}{-1}\addtocounter{Hfootnote}{-1}%
  Raymond A. Yeh\footnotemark, \; Alexander G. Schwing\\
  University of Illinois at Urbana-Champaign\\
  \texttt{\{zr5, yeh17, aschwing\}}@illinois.edu \\
}

\let\cite\citep

\begin{document}
\maketitle
\begin{abstract}
Existing semi-supervised learning (SSL) algorithms use a single weight to balance the loss of labeled and unlabeled examples, \ie, all unlabeled examples are equally weighted. But not all unlabeled data are equal. In this paper we study how to use a different weight for 
\emph{every} unlabeled example. Manual tuning of all those weights 
-- as done in prior work --    is no longer possible. Instead, we adjust those weights via an algorithm based on the influence function, a measure of a model's dependency on one training example.  
To make the approach efficient, 
we propose a fast and effective approximation of the influence function. We demonstrate that this technique outperforms state-of-the-art methods on semi-supervised image and language classification tasks.
\end{abstract}
\vsp{-0.2cm}
\section{Introduction}
Unlabeled data  helps to  reduce the cost of supervised learning, particularly in fields where it is expensive to obtain annotations. 
For instance, labels for biomedical tasks need to be provided by domain experts, which are expensive to hire. Besides the hiring cost,  labeling tasks are often labor intensive, \eg, dense labeling of video data requires  to review many frames.
Hence, a significant amount of effort has been invested to develop novel semi-supervised learning (SSL) algorithms, \ie, algorithms which utilize both labeled and unlabeled data. See the seminal review (specifically Sec.~1.1.2.)~by~\citet{chapelle2006semi} and references therein.
	
Classical semi-supervised techniques~\cite{mclachlan1982updating, shahshahani1994effect,SchwingICML2012,XuCVPR2014} based on expectation-maximization~\cite{Dempster1977, krishnan1997algorithm} iterate between (a) inferring a label-estimate for the unlabeled portion of the data using the current model and (b) using both labels and label-estimates to update the model. Methods for deep nets have also been explored~\citep{Lee2013pseudo, oliver2019benchmark, ren-cvpr2020, ren-eccv2020}. More recently, data augmentation techniques are combined with label-estimation for SSL. The key idea is to improve the model via consistency losses  which encourage labels to remain identical after augmentation~\cite{berthelot2019mixmatch, xie2019uda}.

Formally, the standard SSL setup consists of three datasets: a labeled training set, an unlabeled training set, and a validation set. In practice, SSL algorithms train the model parameters on both the labeled and unlabeled training sets and tune the hyperparameters manually based on the validation set performance. 
Specifically, a key hyperparameter adjusts the trade-off between labeled and unlabeled data. All  aforementioned SSL methods use a \textit{single scalar} for this, \ie, an identical weight is assigned to all unlabeled data points. To obtain good performance, in practice, this weight is carefully tuned using the validation set, 
and changes over the training iterations~\cite{berthelot2019mixmatch}. 


\begin{figure*}
\centering
\setlength{\tabcolsep}{2pt}
\renewcommand{\arraystretch}{0}
\begin{tabular}{ccc}
\includegraphics[width=0.29\textwidth, trim={2.95cm 3.cm 3.8cm 3.2cm},clip]{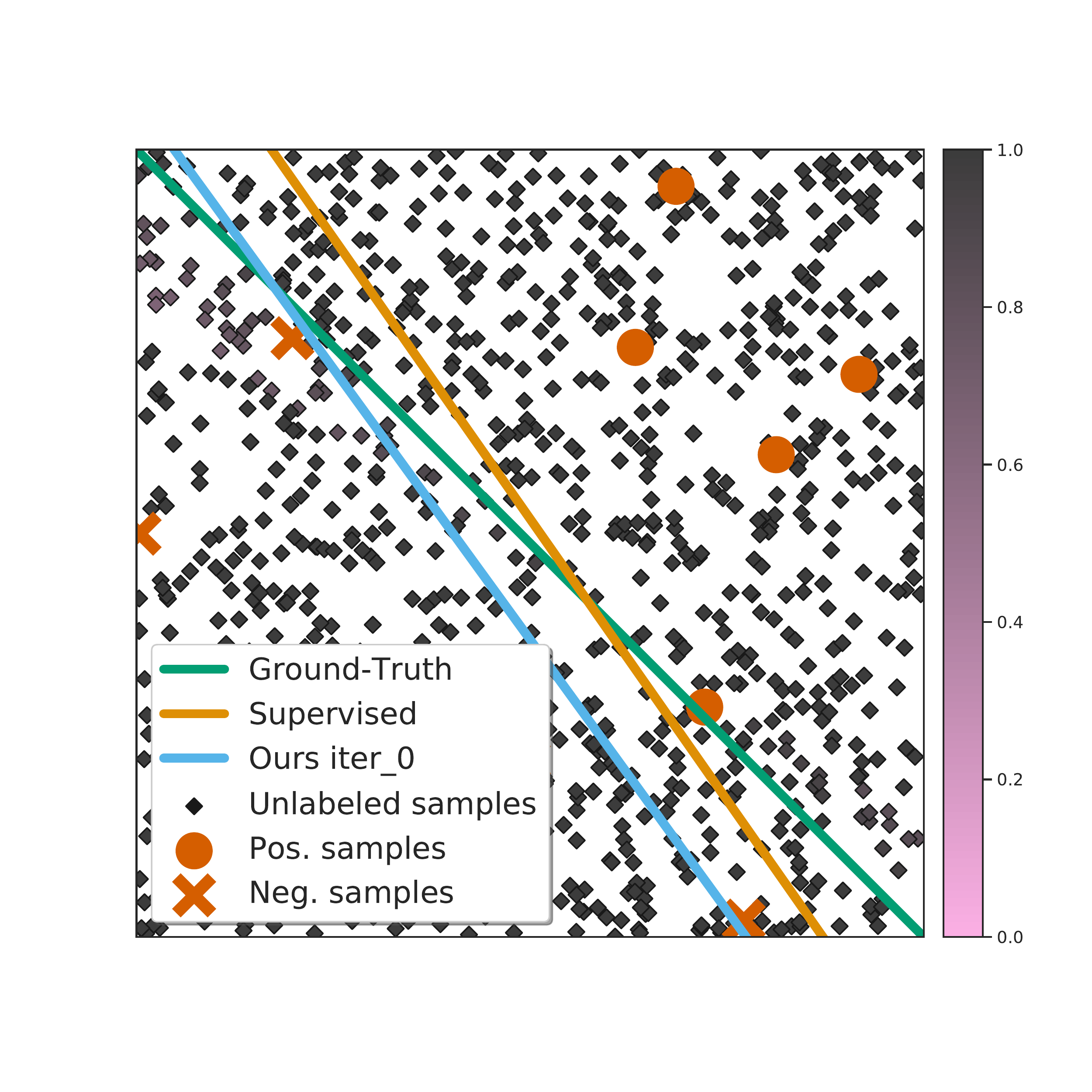}&
\hspace{0.2cm}
\includegraphics[width=0.29\textwidth, trim={2.95cm 3.cm 3.8cm 3.2cm},clip]{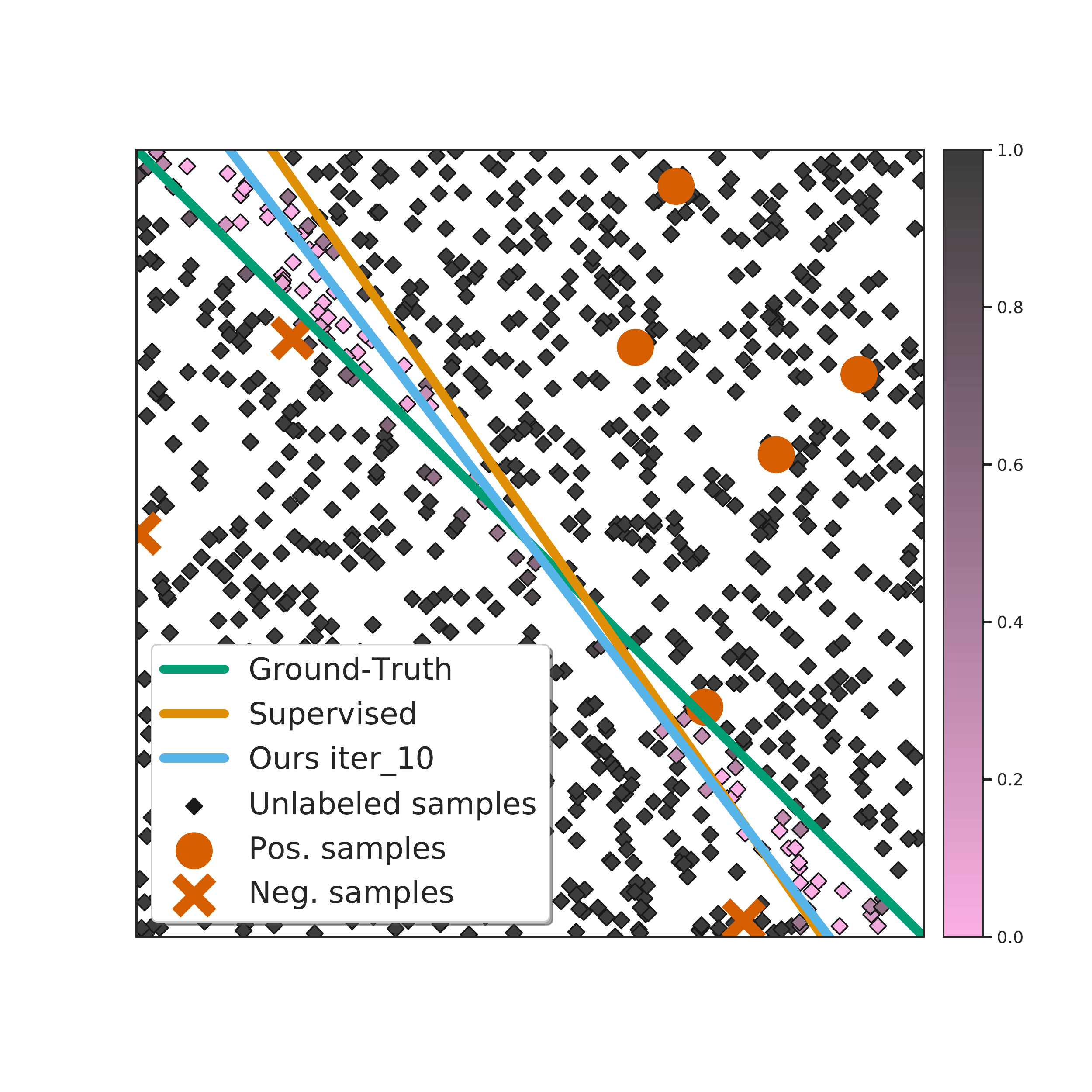} &
\hspace{0.2cm}
\includegraphics[width=0.29\textwidth, trim={2.95cm 3.cm 3.8cm 3.2cm},clip]{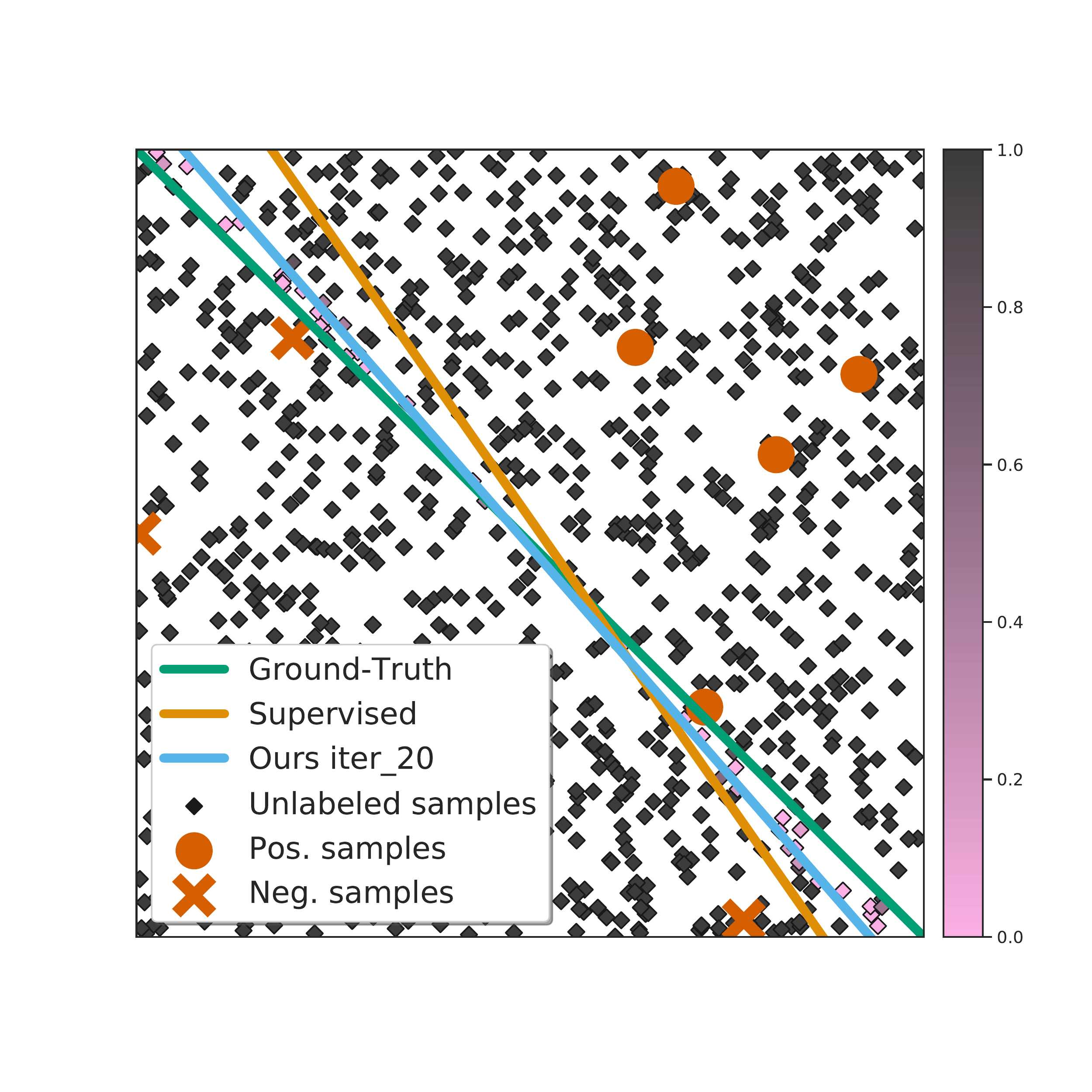}
\includegraphics[width=0.035\textwidth, height=4.2cm, trim={0cm 3.cm 0cm 3.05cm},clip]{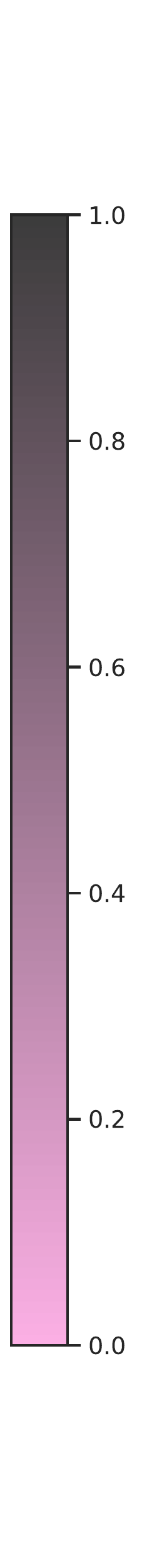} \\
{\footnotesize{Iteration 0}} & {\footnotesize{Iteration 10}} & {\footnotesize{Iteration 20}}
\end{tabular}
\caption{
Decision boundaries across training iterations on linearly separable  data. Labeled samples are shown in {\bf \color{myorange} orange} and unlabeled data in {\bf black}/{\bf \color{mypink}pink} (shading depicts weight of each unlabeled point). 
Our approach {\bf \color{myblue} (blue)} with per example weights with Pseudo label SSL algorithm~\cite{Lee2013pseudo}. 
}
\label{fig:teaser}
\end{figure*}

We think not all unlabeled data are equal. For instance, when the label-estimate of an unlabeled example is incorrect, training on that particular label-estimate  hurts  overall performance. In this case, using a single scalar to weight the labeled and unlabeled data loss term is suboptimal. To address this, we study use of an individual weight \textit{for each} of the unlabeled examples. To facilitate such a large number of hyperparameters, we automatically adjust the per-example weights by utilizing the influence function~\cite{cook1980characterizations}. This influence function estimates the ``importance'' of each unlabeled example using the validation set performance. 

In \figref{fig:teaser} we demo this idea on {\bf \color{myorange} labeled} and {\bf unlabeled},  linearly separable data. The gray/pink color shade indicates the weight of the unlabeled data. We observe the {\bf \color{myblue} proposed method} to more closely mimic  {\bf \color{mygt} ground-truth} compared to {\bf \color{mysp} supervised} training. 


The performance gain does not come for free. The method involves adjusting  per-example weights for each unlabeled example which is computationally expensive if implemented naively. 
Specifically, adjusting a per-example weight involves computing (a) a per-example gradient and (b) an inverse Hessian vector product \wrt the model parameters. To address both challenges, 
we design an efficient algorithm for computing 
per-example gradients, extending backpropagation.
Moreover, we propose an effective and efficiently computable approximation specifically for the influence functions of deep nets. 
These improvements permit to scale the approach to recent SSL tasks and achieve compelling results on  CIFAR-10, SVHN, and IMDb. 

\vsp{-0.1cm}
\section{Background \& Related Work}
\label{sec:related}

We first discusss recent advances in semi-supervised learning for image classification, influence functions and gradient based hyperparameter optimization.

\textbf{Semi-supervised Learning (SSL).}
Given a labeled dataset $\cD=\{(x,y)\}$, a set of unlabeled data $\cU=\{(u)\}$, and a validation set $\cV=\{(x,y)\}$, SSL aims to solve the following program:
\begin{equation}\label{eq:ssl_total}
\min\limits_\theta \sum\limits_{(x,y) \in \cD} \ell_S(x, y, \theta) + 
\lambda \sum\limits_{u \in \cU} \ell_U(u, \theta),
\end{equation}
where 
$\ell_S$ denotes the per-example supervised loss, \eg, cross-entropy for classification, and $\ell_U$ denotes the per-example unsupervised loss, \eg, consistency loss~\cite{xie2019uda} or a regularization term~\cite{Miyato2018vat, berthelot2019mixmatch}.
Lastly, $\theta$ denotes the model parameters and $\lambda \in \mathbb{R}_{\geq 0}$ denotes the scalar weight
which balances the supervised and unsupervised loss terms. Note that existing works use a single non-negative real-valued $\lambda$. Tuning of $\lambda$ is performed either manually or via grid-search based on a performance metric assessed on the validation set $\cV$. 

Different choices of the unsupervised loss $\ell_U$ lead to different SSL algorithms. For example, unsupervised losses $\ell_U(u, \theta)$ resembling a supervised loss $\ell_S$ with the pseudo label $\tilde{y}$, \ie, $\ell_U(u, \theta) \triangleq \ell_S(u, \tilde{y}, \theta)$.
In most cases, the pseudo label $\tilde{y}$ is constructed based on the
model's predicted probability $p_\theta(k|u)$ for class $k$. 
The exact construction of the pseudo label $\tilde{y}$ depends on the  SSL algorithm. 


Specifically, Pseudo-Labeling~\cite{Lee2013pseudo} chooses $\tilde{y}$ to be the  label predicted by the current model $p_\theta(k|u)$, 
\ie, $\tilde{y} = \text{One-Hot}(p_\theta(k|u))$ and uses the cross entropy loss for $\ell_U$. Mean Teacher~\cite{Tarvainen2017meanteacher} chooses $\tilde{y}[k] = \sum_i \alpha^i \cdot p_{\theta_i}(k|u)$ to be an exponential moving average of model predictions, where $\alpha$ is a decay factor and $\theta_i$ denotes the model parameters $i$ iterations ago (0 being the most recent). 
Virtual Adversarial Training (VAT)~\cite{Miyato2018vat}, MixMatch~\cite{berthelot2019mixmatch}, UDA~\cite{xie2019uda}, ReMixMatch~\cite{berthelot2020remixmatch} and FixMatch~\cite{fixmatch} all choose the pseudo-labels based  on predictions of augmented samples, \ie, 
 $\tilde{y}[k] = p_\theta(k|\text{Augment}(u))$.
 
For the augmentation $\text{Augment}(u)$, VAT adversely learns an additive transform, MixMatch considers shifts and image flipping, UDA employs cropping and flipping of the unlabeled images, ReMixMatch learns an augmentation policy during training and FixMatch  uses a combination of augmentations from ReMixMatch and UDA. 
In summary, all these methods encourage consistency under different augmentations of the input, which is imposed by learning with the extracted pseudo-label. 


Note that  all these works use a single scalar weight $\lambda$  to balance  the supervised and unsupervised losses. 
In contrast, we study 
a per-example weight $\lambda_u$ for each $u\in\cU$, as the quality of the pseudo-label varies across unlabeled examples. 

\textbf{Influence Functions.}
Discussed for robust statistics, influence functions measure a model's dependency on a particular training example~\cite{cook1980characterizations}. More specifically, the influence function computes the change $\frac{\partial \theta^*(\epsilon)}{\partial \epsilon}$ of the optimal model parameters when upweighting the loss of a training example $x$ by a factor $\epsilon > 0$, \ie, $\theta^*(\epsilon) \triangleq \arg\min_\theta \sum_{(x', y') \in \cD}\ell_S(x', y') + \epsilon\ell_S(x,y)$.
Recently,~\citet{koh2017understanding}  utilized influence functions to understand black-box models and to perform dataset poisoning attacks. 
Moreover, \citet{koh2019accuracy}  study the accuracy of influence functions when applied 
on a batch of training examples. 
\citet{ren18l2rw} use influence functions in the context of robust supervised learning. 

Different from these works, we develop an influence function based method for SSL. In the context of hyperparameter optimization, influence functions can be viewed as a special case of a hypergradient, where the hyperparameters are the per-example weights $\lambda_u$. We note that this connection wasn't pointed out by prior works. 
A review of gradient based hyperparameter optimization is provided next. 

\textbf{Gradient-based Hyperparameter Optimization.}
Gradient based hyperparameter optimization has been explored for decades~\cite{Larsen, bengio2000gradient, maclaurin2015gradient, luketina2016scalable, Shaban, lorraine2019opt}, and is typically formulated as a bi-level optimization problem: the upper-level and lower-level task maximize the performance on the validation and training set respectively. 
These works differ amongst each other in how the hypergradients are approximated. 
A summary of these approximations is provided in the Appendix~\tabref{tab:hessian}. Theoretical analysis on gradient-based methods for bi-level optimization is also available~\cite{couellan2016convergence, franceschi2018bilevel}. 

In contrast to  existing work which tunes \textit{general hyperparameters} such as weight decay, learning rate, \etc, 
we focus on adjusting the per-example weights in the context of SSL. 
This particular hyperparameter introduces new computational challenges going beyond prior works, \eg, the need for  per-example gradients and  sparse updates. We address these challenges via an efficient algorithm with a low memory footprint and running time. 
Thanks to these improvements, we demonstrate compelling results on semi-supervised image and text classification tasks.

\section{SSL with Per-example Weights}
A drawback of the  SSL frameworks specified in \equref{eq:ssl_total} is their equal weighting of all unlabeled data via a single hyperparameter $\lambda$: 
 all unlabeled samples are treated equally. 
Instead, we study use of a different balance term $\lambda_u \in \mathbb{R}_{\geq 0}$ for each unlabeled datapoint $u\in \cU$. 
This permits to adjust individual samples in a more fine-grained manner. 

However, these  per-example weights introduce a new challenge: manually tuning or grid-search for each $\lambda_u$ is intractable, particularly if the size of the unlabeled dataset is huge. To address this, we develop an algorithm which learns the per-example weights $\lambda_u$ for each  unlabeled data point. Formally, we address the following bi-level optimization problem: 
\begin{align}\label{eq:semi}
& \min_{\Lambda = \{\lambda_1, \ldots, \lambda_{|\cU|}\}} \cL_S(\cV, \theta^*(\Lambda))\;\;  \text{s.t.} \;\; 
\theta^*(\Lambda) = \arg\min_{\theta} \cL_S(\cD, \theta) + \sum_{u\in \cU}\lambda_u \cdot \ell_U(u, \theta),
\end{align}
where $\Lambda  \in \mathbb{R}_{\geq 0}^{|\cU|}$ subsumes  $\lambda_u\; \forall u \in \cU$ and $\cL_S(\cdot, \theta)$ denotes the supervised loss over a labeled dataset, \eg, $\cL_S(\cD, \theta) \triangleq \sum_{(x,y) \in \cD} \ell_S(x, y, \theta)$. Intuitively, the program given in \equref{eq:semi} aims to minimize the supervised loss evaluated on the validation set \wrt the weights of unlabeled samples $\Lambda$, while being given model parameters $\theta^{*}(\Lambda)$ which minimize the overall training loss $\cL(\cD, \cU, \theta, \Lambda) \triangleq \cL_S(\cD, \theta) + \cL_U(\cU, \theta, \Lambda)$. 
Here, $\cL_U(\cU, \theta, \Lambda)$  denotes the weighted unsupervised loss over the unlabeled dataset, \ie, $\cL_U(\cU, \theta, \Lambda) \triangleq \sum_{u\in \cU}\lambda_u \cdot \ell_U(u, \theta)$.

\begin{algorithm}[t]
\begin{algorithmic}[1]
\STATE Initialize model parameters $\theta$, per-example weights $\Lambda$, step size $\eta, \alpha$
\WHILE {not converged} 
\FOR{$1\dots N$}
\STATE Sample batches $\cD' \subseteq \cD$, $\cU' \subseteq \cU$
\STATE $\theta \leftarrow \theta - \alpha \cdot \nabla_\theta \cL(\cD', \cU', \theta, \Lambda)$%
\ENDFOR
\STATE Sample batches $\cD' \subseteq \cD$, $\cU' \subseteq \cU, \cV' \subseteq \cV$
\STATE $\theta^* \leftarrow \theta$
\STATE Compute gradient $\nabla_\theta \cL_{U}(u, \theta, \lambda_u) \;\forall u \in \cU'$ \label{lst:line:per_exp_grad}
\STATE Compute inverse Hessian matrix  $H^{-1}_{\theta^*}$
 \label{lst:line:hessian}
\STATE Approximate $\frac{\partial \cL_S(\cV', \theta^*{(\Lambda)})}{\partial \lambda_u} \;\forall u\in \cU'$ (Eq.~\ref{eq:influence})
\STATE Update per-example weights $\lambda_u \leftarrow \lambda_u - \eta \cdot \frac{\partial \cL_S(\cV', \theta^*(\Lambda))}{\partial \lambda_u} \;\forall u\in \cU'$
\ENDWHILE 
\end{algorithmic}
\caption{SSL per-example weight optimization via influence function.}
\label{alg:training}
\end{algorithm}

%
When optimization involves deep nets and large datasets, adaptive gradient based methods like Stochastic Gradient Descent (SGD) have shown to be very effective time and again~\cite{lecun2015deep, bengio2000gradient}. 
Here too we use gradient based methods for both the inner and outer optimization. Hence, the algorithm iteratively alternates between updating the model parameters $\theta$ and the per-example weights $\Lambda$, as summarized in \algref{alg:training}.
Optimization \wrt $\theta$, while holding $\Lambda$ fixed, involves  several gradient descent updates on the model parameters $\theta$ to reduce the  loss, \ie, 
\be
\theta \leftarrow \theta - \alpha \cdot \nabla_\theta \cL(\cD, \cU, \theta, \Lambda).
\ee
Here, $\alpha > 0$ is the step size. After having updated $\theta$,  $\Lambda$  is adjusted based on the gradient of the validation loss: 
\be
\lambda_u \leftarrow \lambda_u - \eta \cdot \frac{\partial \cL_S(\cV, \theta^*(\lambda))}{\partial \lambda_u} \quad\forall u\in \cU,
\ee
with $\eta>0$ denoting the step size. These two update steps are performed until the validation loss $\cL_S(\cV, \theta(\Lambda))$ converges. 
To compute the updates for $\lambda_u$, we decompose the gradient by applying Danskin's theorem~\cite{danskin1967theory}:
\begin{equation}
\label{eq:implicit}
\frac{\partial \cL_S(\cV, \theta^*(\Lambda))}{\partial \lambda_u} = \nabla_\theta  \cL_S(\cV, \theta^*(\Lambda))^\top \; \frac{\partial \theta^*(\Lambda)}{\partial \lambda_u} \; \forall u \in \cU.
\end{equation}
Recall that $\theta^*$ is a function resulting from an optimization with dependencies on $\Lambda$. Computing the gradient with respect to $\lambda_u$ hence requires differentiating through the optimization procedure or the program $\arg\min_{\theta} \cL_S(\cD, \theta) + \sum_{u\in \cU}\lambda_u \cdot \ell_U(u, \theta)$. Several methods have been proposed to approximate $\frac{\partial \theta^*(\Lambda)}{\partial \lambda_u}$ as discussed in \secref{sec:related}. 

In practice, we found the approximation from~\citet{cook1980characterizations} and~\citet{koh2017understanding} to work well. If $\cL$ is twice differentiable and has an invertible Hessian, then \equref{eq:implicit} can be written as:
\begin{equation}
\label{eq:influence}
\frac{\partial \cL_S(\cV, \theta^*(\Lambda))}{\partial \lambda_u} = -\nabla_\theta  \cL_S(\cV, \theta^*)^\top \;
H_{\theta^*}^{-1} \; \nabla_\theta \ell_{U}(u, \theta^*),
\end{equation}
with the Hessian $H_{\theta^*} \triangleq \nabla_\theta^2 \cL(\cD, \cU, \theta^*, \Lambda)$. Observe that \equref{eq:influence} 
measures how up-weighting a training point changes the validation loss, where the derivative $\frac{\partial \theta^*(\Lambda)}{\partial \lambda_u}$  is approximated using influence functions~\cite{cook1980characterizations}.

When using deep nets, computing~\equref{eq:influence} for all unlabeled examples is challenging. It requires to evaluate per-example gradients for each unlabeled example ($\nabla_\theta \ell_{U}(u, \theta^*)$ $\forall u\in\cU$) and to invert a high dimensional Hessian ($H_{\theta^*}$). Therefore, in the next section, we discuss approximations which we empirically found to be effective when using these techniques for SSL. 

\begin{figure}
\centering
\setlength{\tabcolsep}{2pt}
\renewcommand{\arraystretch}{0}
\begin{tabular}{ccc}
\includegraphics[width=0.29\textwidth, trim={3.1cm 3.3cm 3.9cm 3.4cm}, clip]{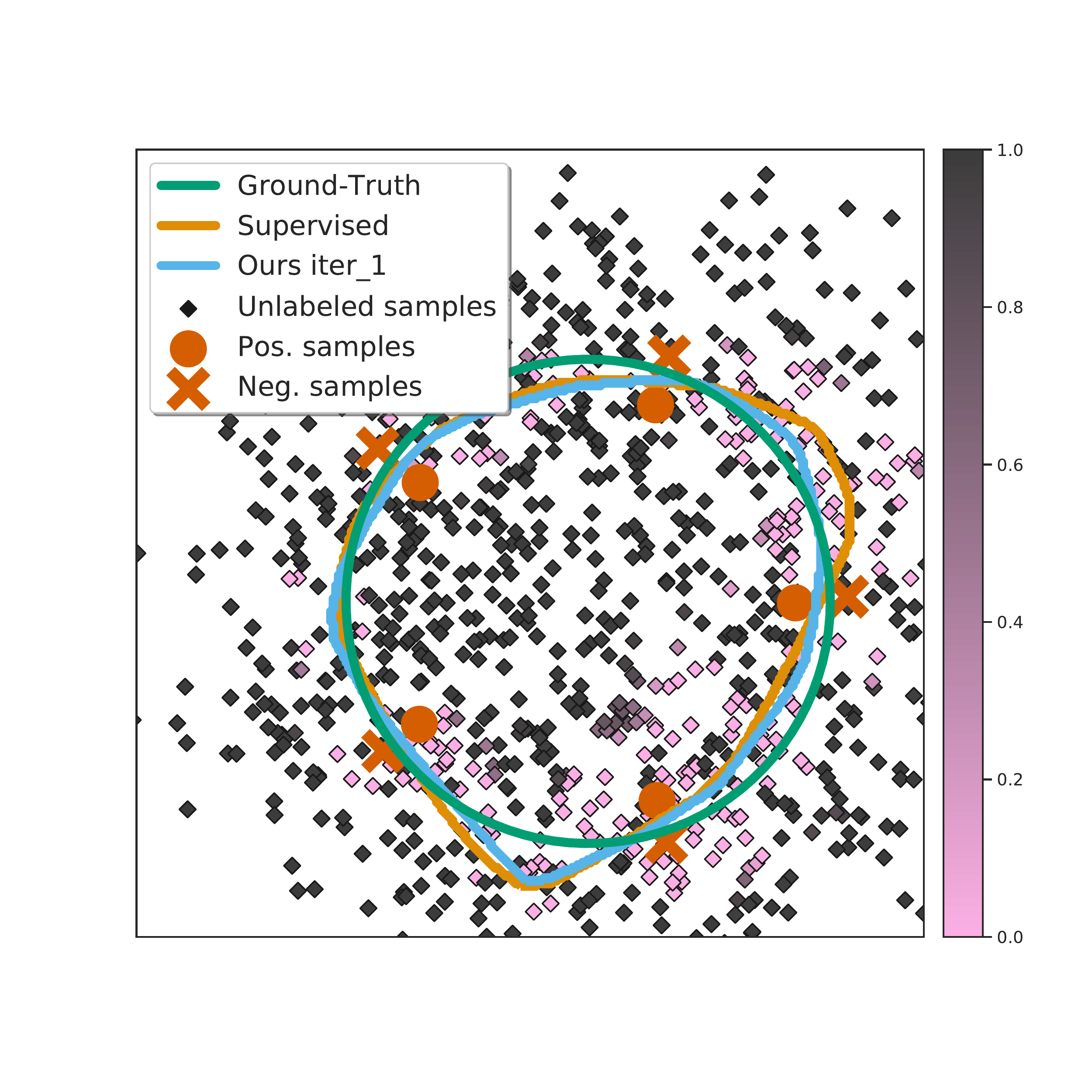}&
\includegraphics[width=0.29\textwidth, trim={3.1cm 3.3cm 3.9cm 3.4cm},clip]{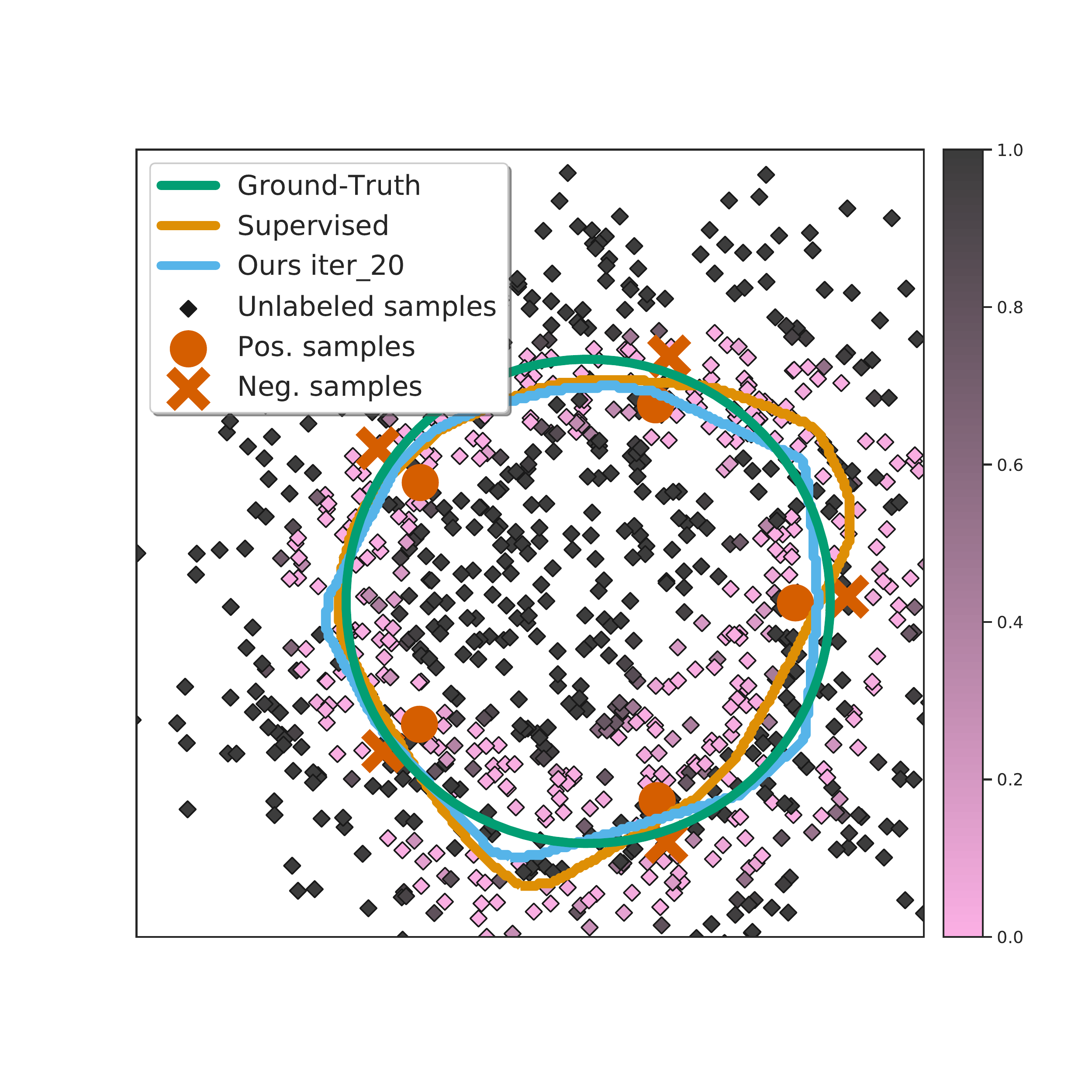}&
\includegraphics[width=0.29\textwidth, trim={3.1cm 3.3cm 3.9cm 3.4cm},clip]{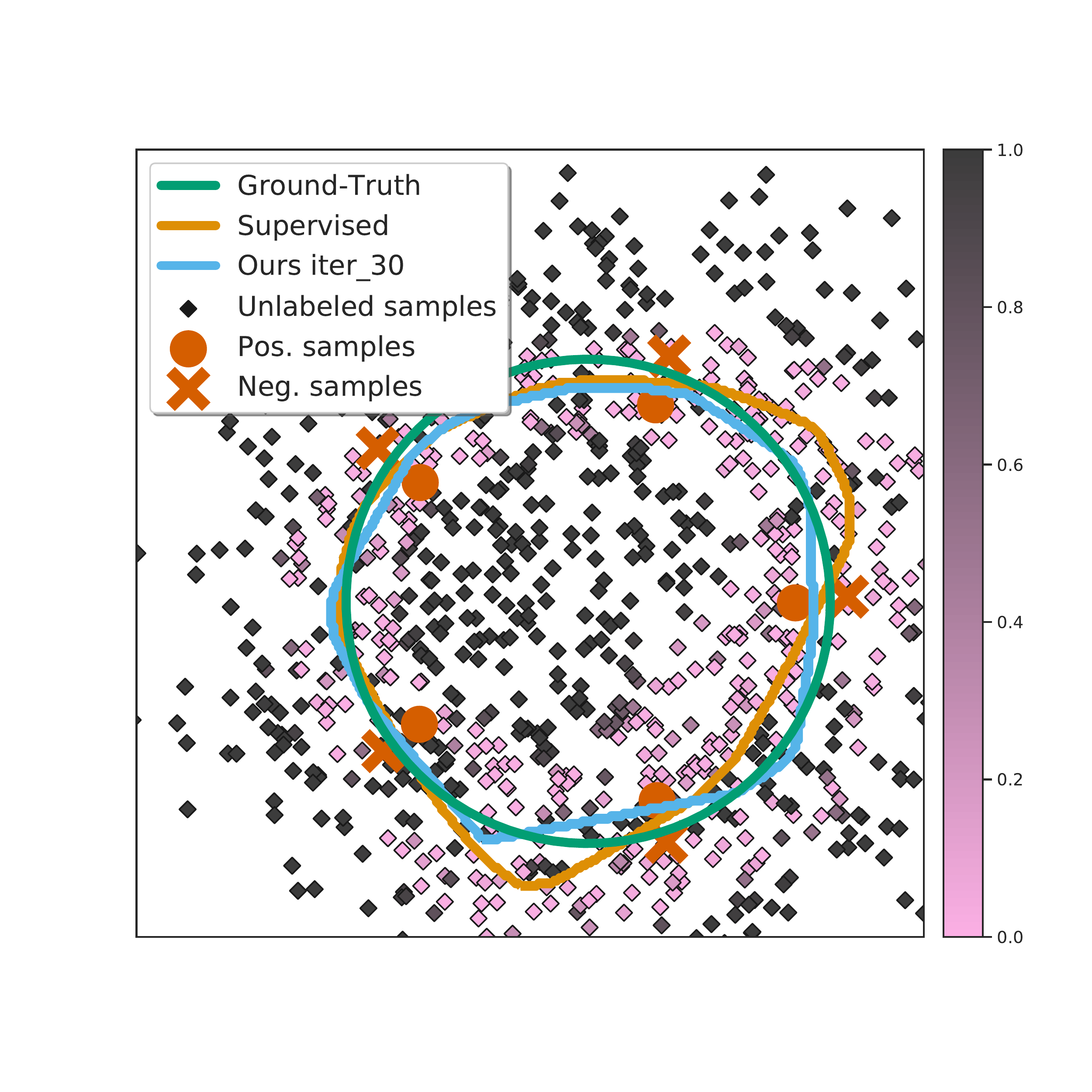}
\includegraphics[width=0.035\textwidth, height=4.2cm, trim={0cm 3.3cm 0cm 2.98cm},clip]{fig/intro_ill/colorbar.pdf}\\
\includegraphics[ width=0.29\textwidth, trim={3.1cm 3.3cm 3.9cm 3.4cm},clip]{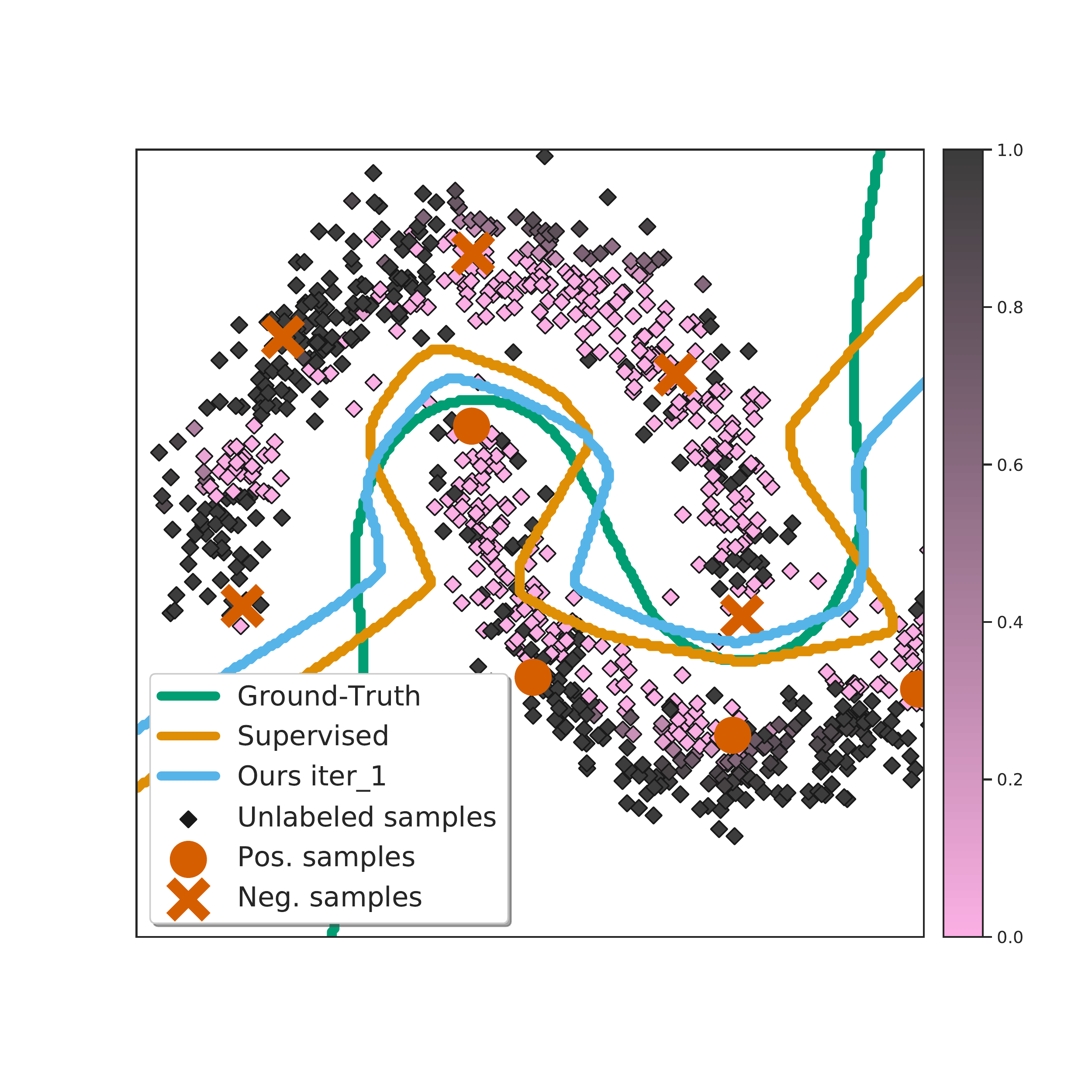}&
\includegraphics[width=0.29\textwidth, trim={3.1cm 3.3cm 3.9cm 3.4cm},clip]{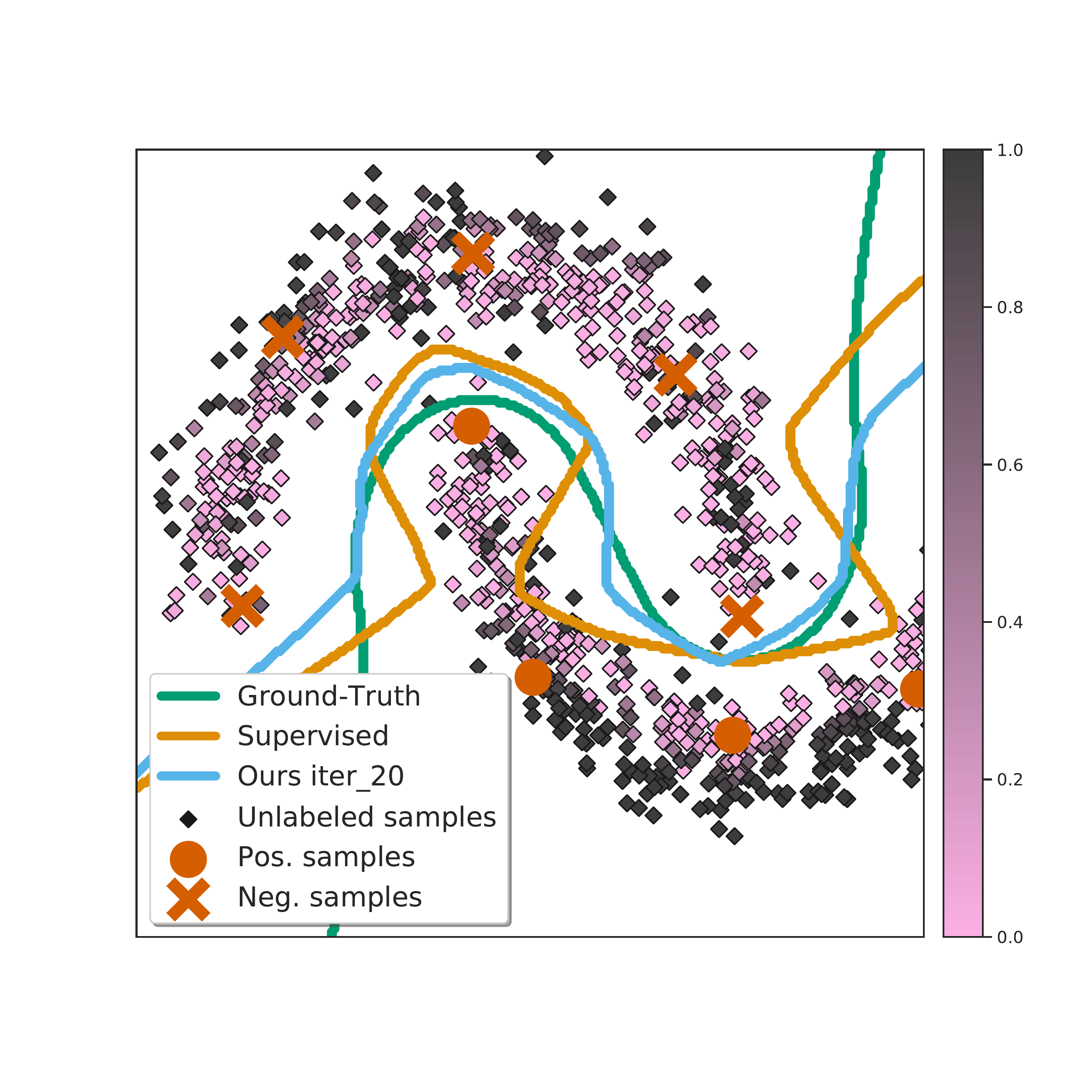}&
\includegraphics[width=0.29\textwidth, trim={3.1cm 3.3cm 3.9cm 3.4cm},clip]{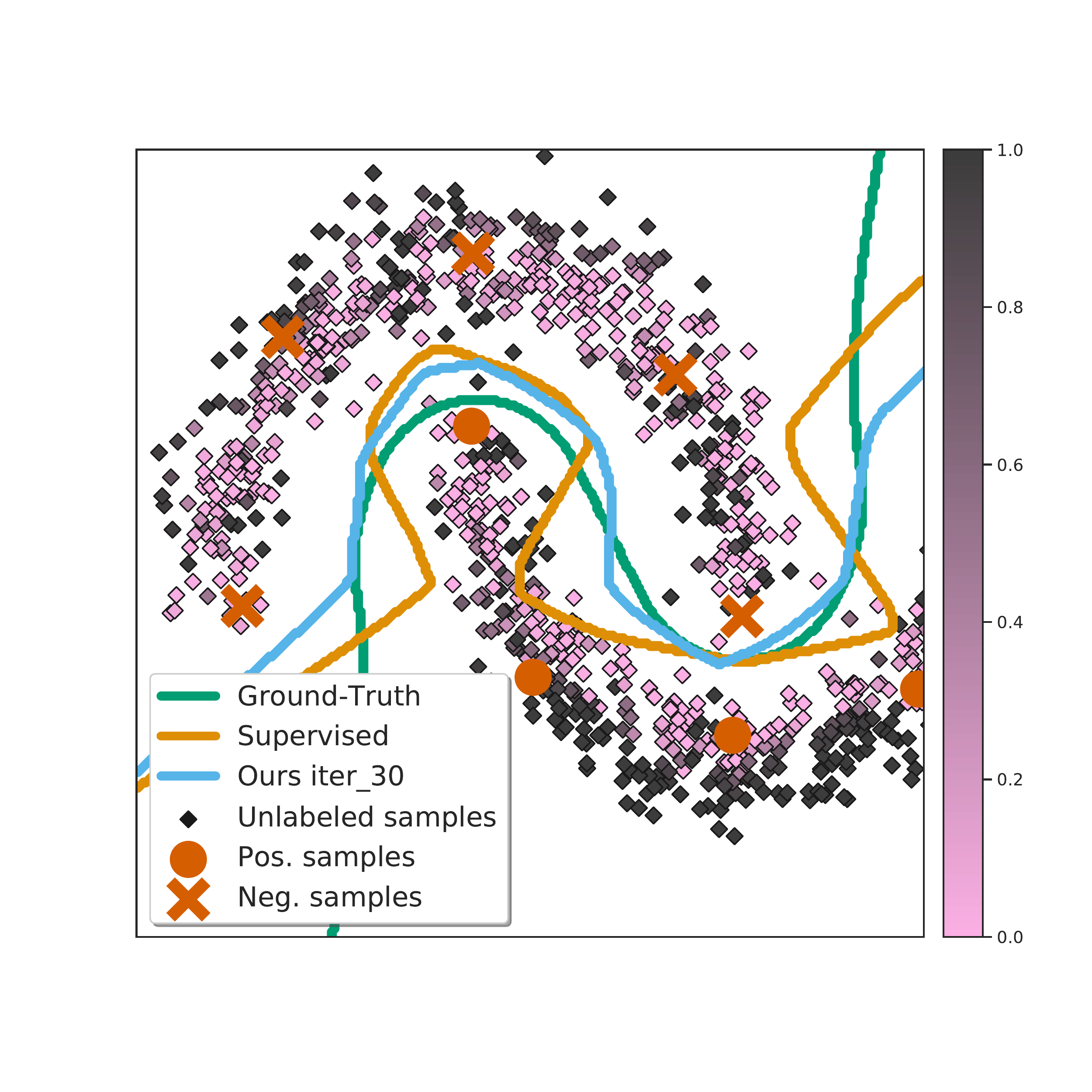}
\includegraphics[width=0.035\textwidth, height=4.2cm, trim={0cm 3.3cm 0cm 2.98cm},clip]{fig/intro_ill/colorbar.pdf}\\
{\footnotesize{Iteration 1}} & {\footnotesize{Iteration 20}} & {\footnotesize{Iteration 30}}
\end{tabular}
\vsp{-.3em}
\caption{The learned decision boundary on the Circles (\textbf{Top}) and Moons (\textbf{Bottom}) dataset. 
Visualization scheme follows~\figref{fig:teaser}.
Observe the changes in weights and the decision boundary. For example, in the top row, the unlabeled examples near the bottom of the circle are down-weighted at iteration 1, which allows for the decision boundary to shrink towards the ground-truth, at iteration 20.}
\label{fig:syn_exp}
\vsp{-0.6cm}
\end{figure}

\subsection{Efficient Computation of Influence Approximation}
As mentioned before, computing the influence function in~\equref{eq:influence} requires addressing two bottlenecks: (a) Computation of per-example gradients (line~\ref{lst:line:per_exp_grad} of~\algref{alg:training}); and (b) Computation of the inverse Hessian (line~\ref{lst:line:hessian} of~\algref{alg:training}). In the remainder of this section, we describe how we tackle both challenges.

{\bf Computation of Per-example Gradient $\nabla_\theta \cL_U(u, \theta)$.} Updating $\Lambda$ requires the gradient of the unsupervised training loss $\cL_U$ \wrt the model parameters $\theta$ individually for each unlabeled point $u \in \cU'$.
However,  backpropagation in deep nets~\cite{rumelhart1986learning}    uses mini-batches and stores cumulative statistics rather than an individual example's gradients. 

A naive solution applies standard backpropagation to mini-batches containing one example, ideally in parallel. 
However, this approach remains too slow for our use case. To improve runtime, we leverage  the fact that standard auto-differentiation tools for deep nets
efficiently compute and store the gradient \wrt a layer \textit{activation} ${h}_u$ for each example $u$.
Applying the chain-rule, the per-example gradient \wrt  the model parameters $\theta$ is then obtained via $\frac{\partial \cL_U}{\partial {h}_u} \cdot \frac{\partial {h}_u}{\partial \theta}$. Hence, we  run standard mini-batch backpropagation to
obtain $\frac{\partial \cL_U}{\partial {h}_u}$ for all  examples in the mini-batch, 
followed by parallel computations which multiply with $\frac{\partial {h_u}}{\partial \theta}$. We describe this approach using  a fully connected layer as an example. 

Consider a per-example loss $\ell_U(u, \theta) \triangleq \ell(\theta^{\top}u)$
with a fully connected layer parametrized by $\theta$. Let $h_u \triangleq \theta^{\intercal}u$ denote the deep net activation for example $u$. 
Auto-differentiation tools compute the gradient \wrt $h_u$ of the loss $\cL_U(\cU', \theta)= \sum_{u \in \cU'} \ell_U(u,\theta)$ over a mini-batch $\cU'$. 
Due to linearity of gradients, $\frac{\partial \cL_U}{\partial h_u} = \frac{\partial \ell_U(u, \theta)}{\partial h_u}$,
which is obtained efficiently for all $u\in \cU'$ in a single backward pass. Next, observe that the per-example gradients \wrt  $\theta$ are efficiently computable on a GPU  via an element-wise multiplication. Note that standard backpropagation employs an inner product as opposed to an element-wise multiplication. 
Information about how to compute per-example gradients for other layers is provided in  Appendix~\ref{sec:supp_per_exp}. 

\textbf{Influence Approximation.}
A second bottleneck for computing the influence function in \equref{eq:influence} is the inverse Hessian $H_{\theta^*}^{-1} $. Directly computing a Hessian  for a modern deep net is not practical due to the huge memory footprint. In addition, computing  its inverse scales worse than quadratically. 
While various approximations have been proposed, they are either too slow or not accurate enough for this  application as we show in~\secref{exp: analysis}. 

Most effective in our study was to approximate~\equref{eq:influence}
by assuming that only the last layer of a deep net
is trainable, \ie, we only consider a subset of the parameters $\hat{\theta} \subset \theta$. Exactly computing the inverse Hessian \wrt $\hat{\theta}$ is reasonably fast as its dimensionality is smaller. 
Importantly, the per-example gradients discussed in the aforementioned paragraph now only need to be computed for $\hat{\theta}$. Consequently, no  backpropagation through the entire deep net is required.  
In~\secref{exp: analysis} we empirically validate that this method greatly accelerates the training process without a loss in accuracy. 

{\bf Efficient Optimizer for $\Lambda$.}
In every iteration the discussed approach updates $\lambda_{u} \; \forall u\in \cU' \subseteq \cU$,
 \ie, only a subset of the weights are considered.
Intuitively, one might implement this by using a separate optimizer for each $\lambda_u$, \ie, a total of $|\cU|$ scalar optimizers. However, this is slow due to the lack of vectorization. To improve, one may consider  a single optimizer for $\Lambda$. However, this approach does not perform the correct computation when the optimizer keeps track of statistics from previous iterations, \eg, momentum. Specificallly, the statistics for all dimensions in $\Lambda$ are updated in every step, even if an example \textit{is not} in the sampled subset, which is not desirable. 

To get the best of both worlds, we modify the latter approach to only update the subset of $\Lambda$ and their statistics that are selected in the subset $\cU'$. We combined this selective update scheme with the Adam optimizer, which we named M(asked)-Adam.
For more details see Appendix~\ref{sec:supp_eff_opt}.
\begin{table*}[t]
\setlength{\tabcolsep}{4pt}
\renewcommand{\arraystretch}{1.05}
\resizebox{\textwidth}{!}{
\centering
\begin{tabular}{c | c c c c c | c c c c c}
\specialrule{.15em}{.05em}{.05em} 
Dataset  &  \multicolumn{5}{c|}{CIFAR-10} &  \multicolumn{5}{c}{SVHN }\\ 
\# Labeled & 250 & 500 & 1000 & 2000 & 4000  & 250 & 500 & 1000  & 2000 & 4000  \\
\hline
Pseudo-Label  & 49.98$\pm$1.17 &40.55$\pm$1.70 &  30.91$\pm$1.73 & 21.96$\pm$0.42 & 16.21$\pm$0.11 &  21.16$\pm$0.88 &14.35$\pm$0.37 & 10.19$\pm$0.41 &7.54$\pm$0.27 & 5.71$\pm$0.07 \\
VAT  & 36.03$\pm$2.82 & 26.11$\pm$1.52 & 18.64$\pm$0.40 &14.40$\pm$0.15 & 11.05$\pm$0.31 & 8.41$\pm$1.01 &7.44$\pm$0.79 & 5.98$\pm$0.21 & 4.85$\pm$0.23 & 4.20$\pm$0.15 \\
Mean-Teacher  & 47.32$\pm$4.71 & 42.01$\pm$5.86 & 17.32$\pm$4.00 & 12.17$\pm$0.22 & 10.36$\pm$0.25 & 6.45$\pm$2.43 &3.82$\pm$0.17 & 3.75$\pm$0.10 & 3.51$\pm$0.09 & 3.39$\pm$0.11  \\
MixMatch  & 11.08$\pm$0.87 &9.65$\pm$0.94 & 7.75$\pm$0.32 &7.03$\pm$0.15 & 6.24$\pm$0.06 & 3.78$\pm$0.26 & 3.64$\pm$0.46 & 3.27$\pm$0.31 &  3.04$\pm$0.13 & 2.89$\pm$0.06 \\
UDA  & 8.76$\pm$0.90 & 6.68$\pm$0.24 &5.87$\pm$0.13 &5.51$\pm$0.21 &5.29$\pm$0.25 & 2.76$\pm$0.17 &2.70$\pm$0.09 &2.55$\pm$0.09 &2.57$\pm$0.09 &2.47$\pm$0.1 5\\
Re-MixMatch  & 6.27$\pm$0.34 & - & 5.73$\pm$0.16 & - & 5.14$\pm$0.04 & 3.10$\pm$0.50 & - & 2.83$\pm$0.30 & - & 2.42$\pm$0.09 \\
FixMatch (CTA) & 5.07$\pm$0.33 & - & - & - & \textbf{4.31$\pm$0.15} & 2.64$\pm$0.64 & - & - & - & 2.36$\pm$0.19 \\
FixMatch* (CTA) & 5.23$\pm$0.28 & - & 4.82$\pm$0.09 & - & 4.48$\pm$0.15 & 2.77$\pm$0.73 & - &  2.41$\pm$0.14 & - & 2.17$\pm$0.08 \\
\hline
Ours (UDA) & \textbf{5.53$\pm$0.17} & \textbf{5.38$\pm$0.23} & \textbf{5.17$\pm$0.16} & \textbf{5.14$\pm$0.17} & \textbf{4.75$\pm$0.28} & \textbf{2.45$\pm$0.08} & \textbf{2.39$\pm$0.04} & \textbf{2.33$\pm$0.06}  & \textbf{2.32$\pm$0.06}  & \textbf{2.35$\pm$0.05} \\
Ours (FixMatch, CTA) & \textbf{5.05$\pm$0.12} & - & \textbf{4.68$\pm$0.14} & - & \textbf{4.35$\pm$0.06} & \textbf{2.63$\pm$0.23} & - &  \textbf{2.34$\pm$0.15} & - & \textbf{2.15$\pm$0.03} \\
\specialrule{.15em}{.05em}{.05em} 
\end{tabular}}
\caption{Test error rate (\%) of methods using Wide ResNet-28-2 on CIFAR-10 and SVHN. For our method, we report the mean and standard deviation over 5 runs. (*: reproduced using released code.)}
\label{tab:res1}
\end{table*}
\vsp{-0.1cm}
\section{Experiments}
In this section, we first analyze the effectiveness of our method on low-dimensional datasets before evaluating on standard SSL benchmarks including CIFAR-10~\cite{krizhevsky2009learning}, SVHN~\cite{netzer2011reading}, and IMDb~\citep{maas-EtAl}. The method achieves compelling results on all benchmarks. Finally, we ablate  different components of the method to illustrate  robustness and efficiency. 
For implementation details, please refer to Appendix~\ref{sec:supp_implement}.

\subsection{Synthetic Experiments}
{\bf Datasets and Model.} Beyond the linearly separable data shown in~\figref{fig:teaser}, we consider two additional datasets with non-linear decision boundary, Circles and Moons.
The Circle dataset's decision boundary  forms a circle, and the Moon dataset's decision boundary has the shape of two half moons, as shown in~\figref{fig:syn_exp}. Each dataset consists of 10 labeled samples, 30 validation examples\footnote{
In  SSL literature, the validation set is commonly larger than the training set, \eg, prior works use 5k validation data when there are only 250 labeled samples~\cite{oliver2019benchmark}.} 
and 1000 unlabeled examples.
We train a deep net consisting of two fully-connected layers with 100 hidden units followed by a ReLU non-linearity. The models are trained following~\algref{alg:training} using Adam optimizer and using pseudo label~\cite{Lee2013pseudo} as the base SSL algorithm. 

{\bf Discussion.} The  approach successfully learns models that fit the ground-truth decision boundary on both datasets. As illustrated using colors in~\figref{fig:syn_exp},  unlabeled examples that are down-weighted the most are near but on the wrong side of the learned decision boundary. This demonstrates that the influence function successfully captures a model's dependency on the training examples. By adjusting the per-example weights on the unlabeled data, the model was able to more closely match the ground-truth. 

\subsection{Semi-supervised Learning Benchmarks}\label{subsec:ssl_benchmark}
We now evaluate our method using per-sample weights on $\ell_U$ defined by UDA~\citep{xie2019uda} and FixMatch~\citep{fixmatch}.

\textbf{Image Classification.} 
Experiments are conducted on CIFAR-10 and SVHN and results are compared to recent methods including Pseudo-Label~\citep{Lee2013pseudo}, VAT~\citep{Miyato2018vat}, Mean-Teacher~\citep{Tarvainen2017meanteacher}, MixMatch~\citep{berthelot2019mixmatch}, UDA~\citep{xie2019uda}, ReMixMatch~\citep{berthelot2020remixmatch}, and FixMatch~\cite{fixmatch}. Following these works, we use Wide-ResNet-28-2~\cite{Zagoruyko2016WRN} with 1.5M parameters  for all experiments for a fair comparison.

We experiment with a varying number of labeled examples from 250 to 4000 and a validation set of size 1024. For completeness we provide in~\secref{exp: analysis} an ablation w.r.t.\ different validation set sizes,  from 64 to 5000. Note that the validation set is \textit{smaller} than that of prior works: MixMatch, Re-MixMatch, and FixMatch use a validation set size of 5000, as specified in their released code. Pseudo-Label, Mean-Teacher, and VAT use a size of 5000 for CIFAR10 and 7000 for SVHN (see~\citet{oliver2019benchmark}). 
We use a smaller validation set as we think 5000 validation examples isn't a practical amount: a setting with 250 labeled training samples  would result in $20\times$ more validation samples. 

SSL benchmark results are provided in Tab.~\ref{tab:res1}. Observe that across different splits the best model outperforms all prior methods achieving improvements over recent baselines like UDA and FixMatch. {\bf For UDA:} the method outperforms the UDA baseline across all splits in both CIFAR-10 and SVHN. {\bf For FixMatch:} we use their best variant of CTAugment and report the numbers from the original paper~\cite{berthelot2020remixmatch} (See FixMatch (CTA) in Tab.~\ref{tab:res1}). To reproduce the numbers (FixMatch* (CTA) in Tab.~\ref{tab:res1}) we use the released code which seems to result in numbers that differ slightly. Observe that per-example weighting is able to  improve upon the original FixMatch baseline results over all splits. 

\begin{table}[t]
\centering
\footnotesize{
\begin{tabular}{c c | c  c}
\specialrule{.15em}{.05em}{.05em} 
Max seq. length & \# Labeled & Methods & Error \\
\hline
no truncation & 25,000 & \citet{Dai_NIPS2015} & 7.24\\
400 & 25,000 & \citet{Miyato2018vat} & 5.91 \\
512 & 25,000 & BERT~\citep{devlin2018bert} & 4.51  \\
no truncation & 25,000 & \citet{SachanZS19} & 4.32  \\
512 & 20 & UDA~\citep{xie2019uda}  & 4.2 \\
\hline
128 & 20 & Supervised & 39.40 \\
128 & 20 & UDA~\citep{xie2019uda} & 8.98$\pm$0.26 \\
128 & 20 & Ours & \textbf{ 8.51$\pm$0.14} \\
\specialrule{.15em}{.05em}{.05em} 
\end{tabular}}
\caption{IMDb classification test error rate (\%). We report the mean and standard deviation over 3 runs for UDA and our method.}
\label{tab:res-imdb}
\end{table}
\textbf{Text Classification.}
We further evaluate the method on  language domain data using the IMDb dataset for binary polarity classification. IMDb consist of $25k$ movie reviews for training data and $25k$ for testing. This dataset comes with $50k$ additional unlabeled data and is therefore widely used to evaluate SSL algorithms. 

Following the  experimental  setup of UDA,  the model is initialized using parameters from BERT~\cite{devlin2018bert}  and fine-tuned on IMDb. We use 20 labeled samples for the supervised training set and another 20 for validation. The remaining data is treated as unlabeled. 

Note that the maximum sequence length is an important factor in determining the final performance. Normally, the longer the sequence, the better the results. The best result of UDA is achieved using a length of 512 on v3-32 Cloud TPU Pods. However, we mostly have access to 16GB GPUs and very limited access to 32GB GPUs. Due to this hardware constraint, we report results with a maximum sequence length of 128.

The results are shown in Tab.~\ref{tab:res-imdb}, where per-example weights achieve a performance gain over the UDA baseline in the 128 max sequence length setting. For completeness we provide results, with various max sequence lengths from recent SSL approaches in the top half of~\tabref{tab:res-imdb}. 

\subsection{Ablation Studies and Analysis}
\label{exp: analysis}
In this section, we perform numerous ablation studies to confirm the efficacy for each of the  components. All the experiments are conducted using CIFAR-10 and the UDA baseline.

{\bf Comparison of Influence Function Approximation.}
We compare the method with recent Hessian approximations: \citet{luketina2016scalable}  approximate the inverse hessian using an identity matrix, and \citet{lorraine2019opt} use the Neumann inverse approximation for efficient computation. 
Note that for Wide-ResNet-28-2 the Neumann approximation requires a large memory footprint  as  recent SSL algorithms use large batch sizes during training. With a 16GB GPU, we are unable to apply their approximation to all the model parameters. To address this, we only apply their approach to the last ResNet block and to the classification layers. 
 
\begin{wrapfigure}{r}{5.5cm}
\centering
\vsp{-0.3cm}
\includegraphics[width=0.95\linewidth, trim={0 0.9cm 0 0.75cm},clip]{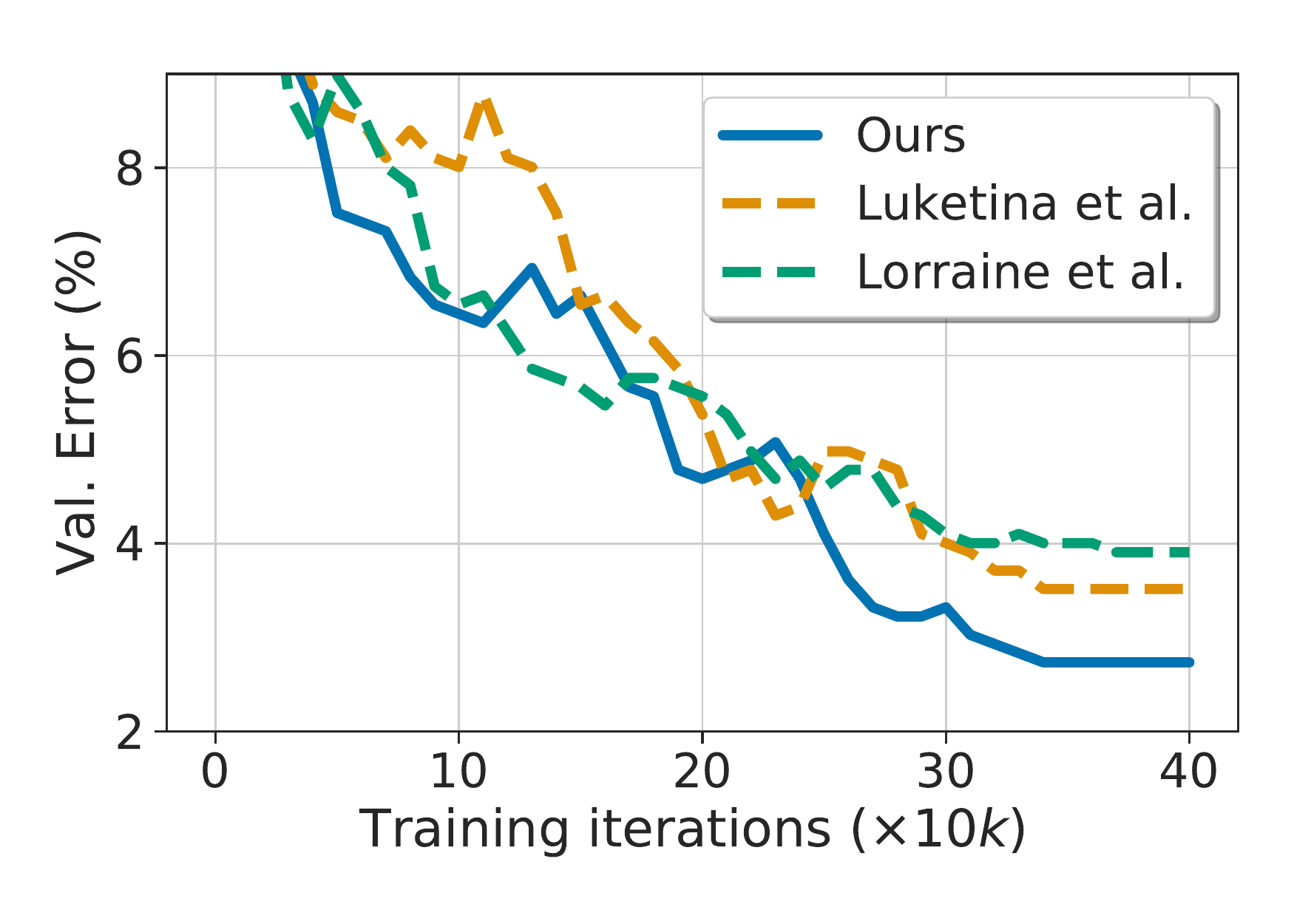}
\vspace{-1em}
\caption{Val. error rate (\%) over training iterations  for different approximations of the influence function.}
\label{fig:ihvp}
\vspace{-0.3cm}
\end{wrapfigure}

In~\figref{fig:ihvp}, we plot the validation error rate over training iterations. In the earlier iterations, the method 
 is on par with~\citet{lorraine2019opt}. In the final iterations, the  approach outperforms the baselines. We suspect that the earlier layers in the model have converged, hence, computing the influence based on the exact inverse Hessian of the last layer becomes accurate. In contrast, baselines will continue to compute the influence based on an approximated inverse Hessian. Hence use of the exact inverse leads to better convergence. The improvement on validation performance also transfers to the test set. 
Ultimately, the method achieves a test error of 4.43\%, outperforming 4.51\% and 4.85\% by \citet{luketina2016scalable} and \citet{lorraine2019opt}, respectively. 

\begin{figure*}[t]
\centering
\includegraphics[height=3.cm, width=0.32\textwidth, trim={0 0.9cm 0 0.75cm},clip]{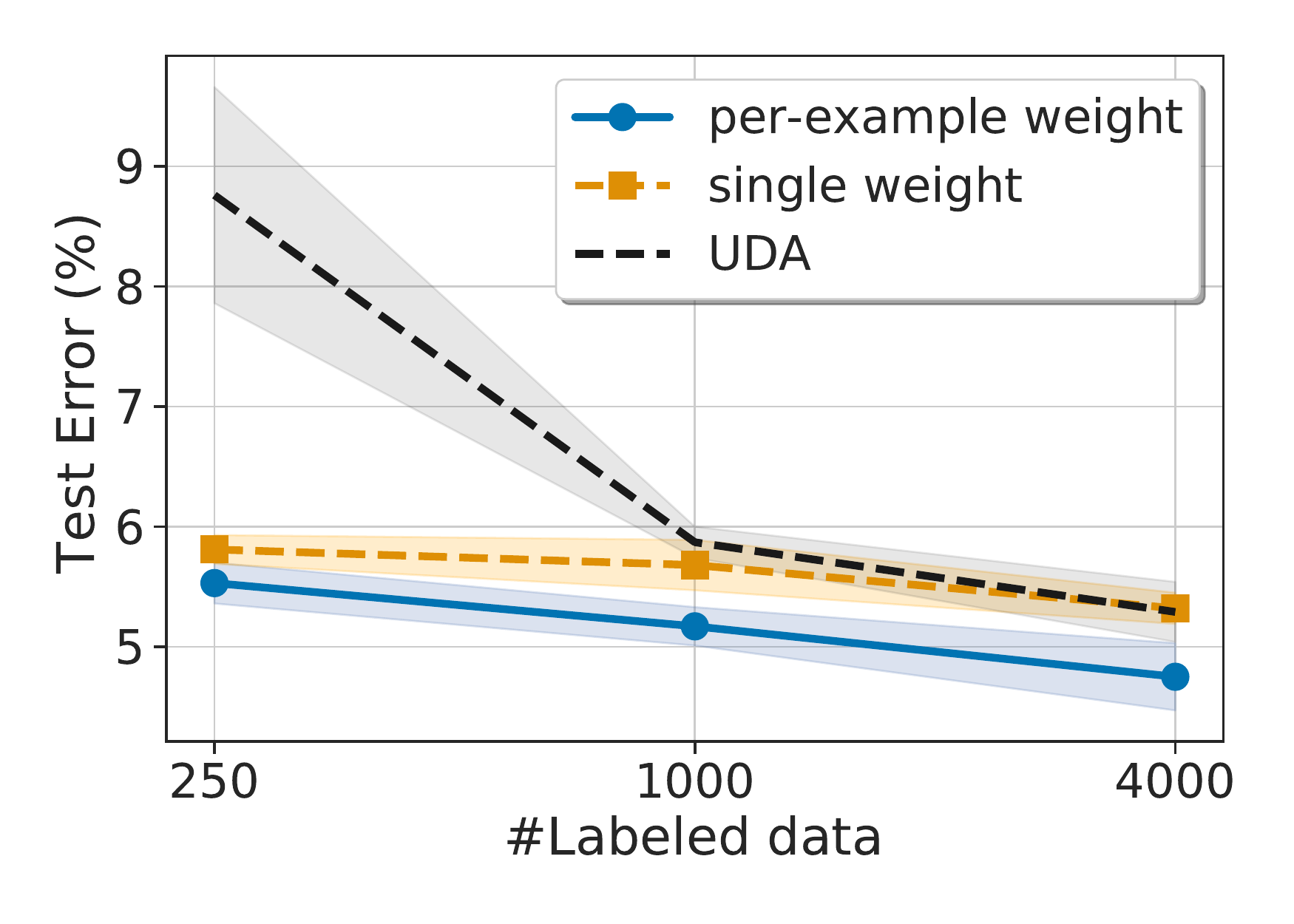}
\includegraphics[height=3.cm, width=0.32\textwidth, trim={0 0.9cm 0 0.75cm},clip]{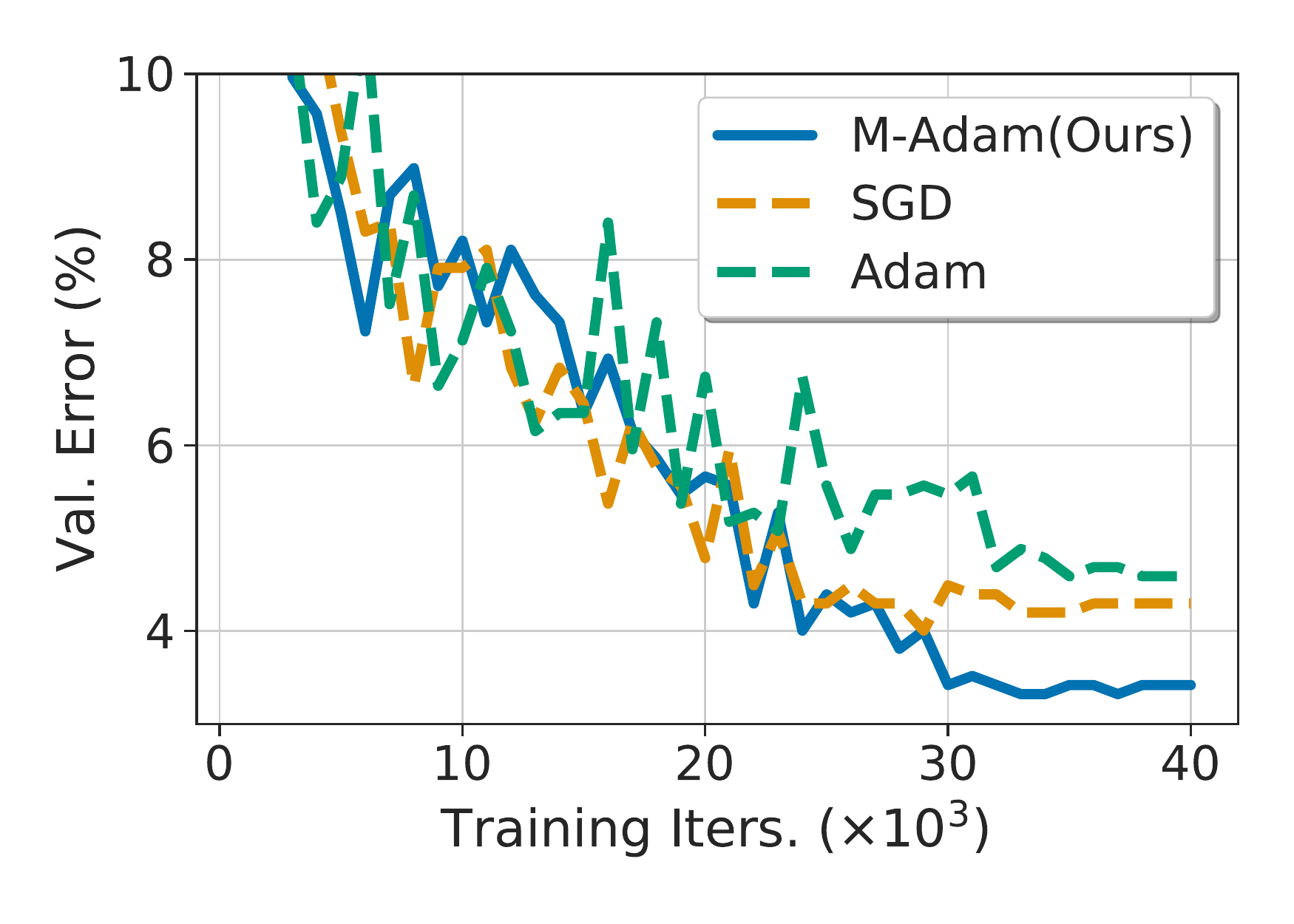}
\includegraphics[height=3.cm, width=0.32\textwidth, trim={0 0.9cm 0 0.75cm},clip]{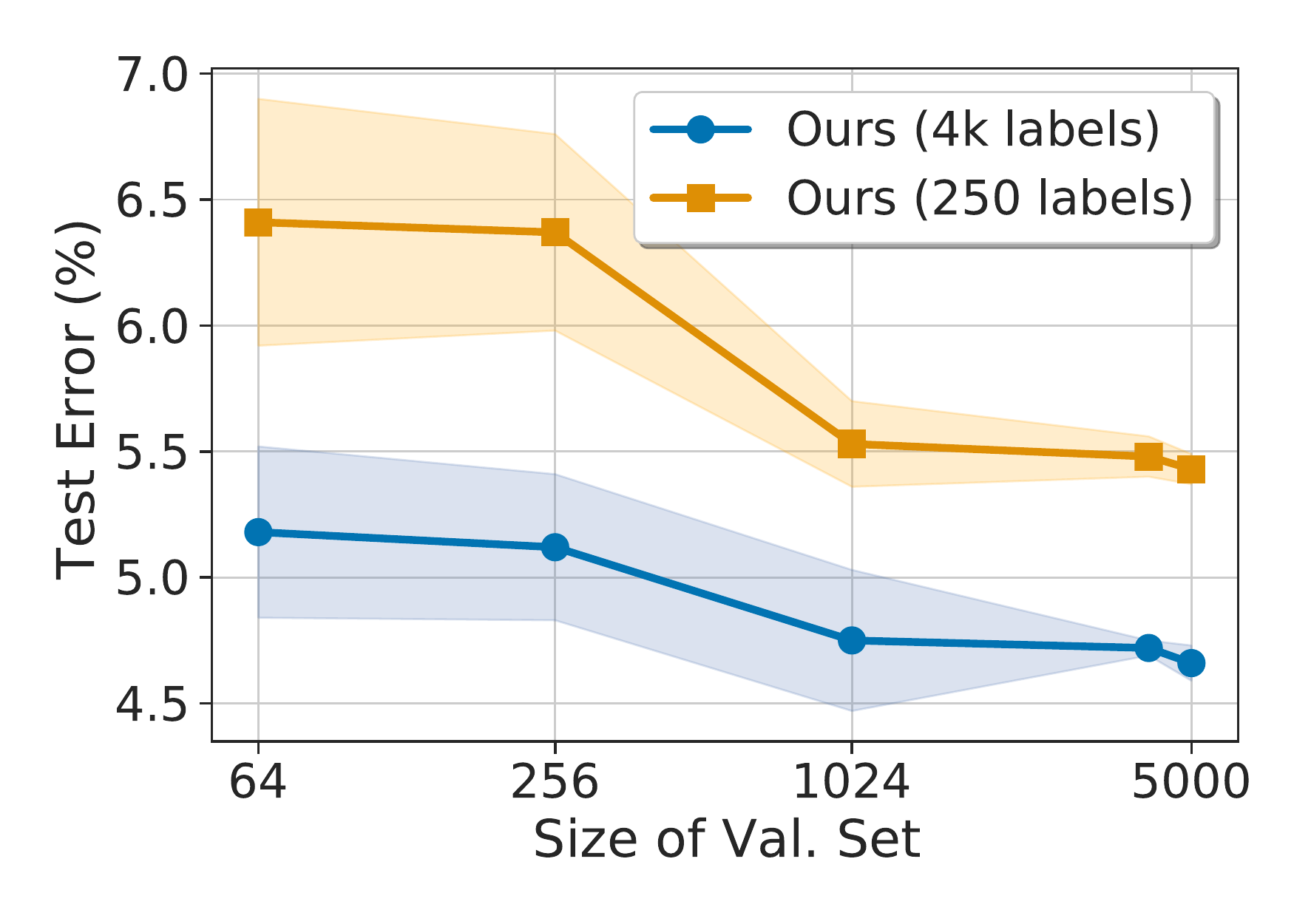}
\caption{
\textbf{Left:} Test error comparison between tuning a single weight and per-example weights over different amounts of labeled data. \textbf{Center:} Validation error during training of models using different optimizers. \textbf{Right:} Test error comparison of models using different validation set sizes. All experiments are conducted on CIFAR-10.
}
\label{fig:abl-three}
\end{figure*}

{\bf Tuning a Single Weight $\lambda$.} To demonstrate the benefits of  per-example weights, we perform an ablation study isolating this factor. We apply the method to tuning of a single $\lambda$, shared across all unlabeled examples, following~\equref{eq:ssl_total}.
As shown in \figref{fig:abl-three} (left), models with per-example weights  outperform models with a single $\lambda$  across different data splits. This verifies the hypothesis that not all unlabeled data are equal and that this method can adjust these weights effectively to improve model performance. Average results over three runs are reported.

{\bf Ablation on Adam Implementation.}
We demonstrate the effectiveness of the  M(asked)-Adam in~\figref{fig:abl-three} (center). We compare with vanilla Adam and SGD.  
We observe that  M-Adam performs the best, followed by SGD, and lastly  vanilla Adam. This result highlights the importance of  masked updates to correctly compute the running averages of gradient statistics. 

{\bf Effect of Validation Size.}
As for all SSL algorithms, the validation set size plays an important role for performance. We study the effect of the validation set size on the final performance when using the proposed method. As shown in \figref{fig:abl-three} (right), results improve consistently from a small validation set (64 samples) to a relatively large one (5000 samples) for both 250 and 4000 labeled data. Average results over three runs are reported.

{\bf Robustness to Hyperparameters}
\algref{alg:training} introduces two  hyperparameters: the inner steps $N$ and the step size $\eta$ for tuning $\Lambda$. We study the robustness to these hyperparameters following the UDA setup. Results are shown in Tab.~\ref{tab:abl-hp}. We observe that a large or small  $N$  hurts the overall performance. 
Similarly, the step size $\eta$ for updating $\Lambda$ in the outer loop of~\algref{alg:training} affects the balance between the two updates for $\theta$ and $\lambda$. We found that the sweet spot is reached at  ($N=100, \eta=0.01$) for CIFAR-10 with 4000 labeled data. We use  these hyperparameter values for 
all splits across the CIFAR-10 and SVHN datasets and found them to work well.

\begin{table}[t]
\centering
\footnotesize{
\begin{tabular}{c | c c |c| c c }
\specialrule{.15em}{.05em}{.05em} 
$(N, \eta)$ & $(30, 0.01)$ & $(300, 0.01)$ & $\bf{(100, 0.01)}$ & $(100, 0.1)$ & $(100, 0.001)$ \\
\hline
Err. & 5.13 & 4.59 & \textbf{3.42} & 6.16 & 4.10 \\
\specialrule{.15em}{.05em}{.05em} 
\end{tabular}
\caption{Ablation study on hyperparameters $N, \eta$. We report the val. error rates on CIFAR-10 with 4000 labeled data.}
\label{tab:abl-hp}}
\end{table}

\subsection{Running Time Comparisons}
\label{sec:speed}
We provide running time results using Wide-ResNet-28-2 with  a batch size of 64, 256, 320 for labeled, unlabeled and validation data respectively. We report the mean running time over 20 iterations.

\textbf{Per-example Gradient.}
We consider  two baseline implementations for computing per-example gradients: a \textit{serial} implementation which iterates over each example in the batch, and a \textit{parallel} implementation using \texttt{tf.vectorized$\_$map}. The serial implementation requires $18.17$s on average for a batch of unlabeled examples to compute the gradients for the entire model. Our method achieves $0.94$s, which is $19.3\times$ faster. The \textit{parallel} implementation requires a much larger memory footprint and no longer fits into a 16GB GPU.

\textbf{Influence Approximation.}
We compare our approximation's running time with~\citet{luketina2016scalable} and \citet{lorraine2019opt}. 
Our approximation takes $0.455$s per batch on average with exact inverse Hessian of the classifier layer, which is comparable to work by \citet{luketina2016scalable} ($0.399$s) which use an identity matrix as the inverse Hessian. 
Note that we implemented \citet{luketina2016scalable}'s approximation using our fast per-example gradient implementation, which again verifies its effectiveness and general utility. 

When compared to \citet{lorraine2019opt}, the approach is $4.6\times$  faster. Their method iteratively approximates the inverse Hessian vector product. Due to the aforementioned (\secref{exp: analysis}) GPU memory constraint,~\citet{lorraine2019opt}'s approach is implemented only on the last ResNet block and the classification layer, which uses 15.8GB of GPU memory. In contrast, the GPU memory consumption of our approach is only 9GB.

\section{Conclusion}
We demonstrate that use of a per-example weight for each unlabeled example helps to improve existing SSL techniques. In contrast to manual tuning of a single weight for all unlabeled examples, as done in prior work, we study an algorithm which automatically tunes these per-example weights through the use of influence functions. 
For this, we develop  solutions to address the computational bottlenecks when computing the influence functions, \ie, the influence approximation and the per-example gradient computation. These improvements permit to scale  to realistic SSL settings and to achieve compelling results on semi-supervised image and text classification benchmarks. 
\clearpage
\section*{Broader Impact}

We propose a method to improve existing semi-supervised learning (SSL) techniques, \ie, achieving better model performance using a limited amount of labeled data. In general, SSL has a large impact on machine learning applications where labeled data are not widely available, \eg, biomedical data, or applications where labeling is expensive, \eg, dense labeling of videos. While our research focuses on classification  benchmarks for SSL, in general, improving SSL techniques will further broaden the scope which machine learning can be applied to. 

Due to this we  foresee a potential positive social impact from our work. In general, we observe that data are being labeled based on the demand of the users. Consider speech recognition datasets: for  common  languages large scale corpora exists, \eg, the LibriSpeech ASR corpus~\cite{panayotov2015librispeech} contains over 1000 hours of English speech. However, very few datasets exist for rare dialects. 

In other words, minority groups may benefit less from  progress in machine learning as the datasets are not collected/labeled. We hope that improvements in SSL will make machine learning more accessible and applicable to everyone as it reduces the need for a collection of large scale labeled data. 

\begin{ack}
This work is supported in part by NSF under Grant No.\ 1718221, 2008387 and MRI \#1725729, NIFA award 2020-67021-32799, UIUC, Samsung, Amazon, 3M, and Cisco Systems Inc.\ (Gift Award CG 1377144). We thank Cisco for access to the Arcetri cluster. We thank Amazon for EC2 credits.  RY is supported by a Google PhD Fellowship. ZR is supported by Yunni \& Maxine Pao Memorial Fellowship. 
\end{ack}
\clearpage

{\small
\bibliographystyle{abbrvnat}
\bibliography{auto_trust}
}

\clearpage
\appendix
\renewcommand{\thetable}{A\arabic{table}}
\setcounter{table}{0}
\setcounter{figure}{0}
\renewcommand{\thetable}{A\arabic{table}}
\renewcommand\thefigure{A\arabic{figure}}
\renewcommand{\theHtable}{A.Tab.\arabic{table}}
\renewcommand{\theHfigure}{A.Abb.\arabic{figure}}
\renewcommand\theequation{A\arabic{equation}}
\renewcommand{\theHequation}{A.Abb.\arabic{equation}}

\newcommand*{\dictchar}[1]{
    \clearpage
    \twocolumn[
    \centerline{\parbox[c][3cm][c]{\textwidth}{
            \centering
            \fontsize{14}{14}
            \selectfont
            {#1}}}]
}

\onecolumn
{\centering \Large \textbf{Appendix}}

In this appendix we first provide additional background (\secref{sec:supp_background}) before detailing more information on per-example gradient computation (\secref{sec:supp_per_exp})  and optimizer implementation (\secref{sec:supp_eff_opt}). 
We then provide 
implementation details (\secref{sec:supp_implement}) and more information about influence functions (\secref{sec:supp_inf_func}). %

\section{Additional Background}
\label{sec:supp_background}
\subsection{Gradient-based Hyperparameter Optimization}%
\begin{table}[H]
\resizebox{\textwidth}{!}{
\centering
\begin{tabular}{c|c|c}
\specialrule{.15em}{.05em}{.05em}
Larsen \etal~\cite{Larsen} & Conjugate gradients (CG)~\cite{Martens}  & Identity~\cite{luketina2016scalable}\\
$\nabla_\theta \cL_S(\cV) \left[ \frac{\partial \cL}{\partial \theta} \frac{\partial \cL^\top}{\partial \theta} \right]^{-1}$
&$ \arg \min_x \|x H_{\theta} -\nabla_\theta \cL_S(\cV) \| $ 
&$\nabla_\theta \cL_S(\cV) \left[ I \right]^{-1} $\\
\hline
\hline
Stochastic CG~\cite{koh2017understanding} & Truncated Unrolled Diff.~\cite{Shaban}  & Neumann~\cite{lorraine2019opt}\\
Using~\cite{agarwal2017second}
& $\nabla_\theta \cL_S(\cV)  \sum_{L<j<i} \left[ \prod_{k<j} I-H_{\theta}|w_{i-k}\right] $
& $\nabla_\theta \cL_S(\cV) \sum_{j<i}\left[ I - \frac{\partial \cL^2_T}{\partial \theta \partial \theta^\top}  \right]^j$\\

\specialrule{.15em}{.05em}{.05em} 
\end{tabular}
}
\caption{A summary of methods to approximate the inverse Hessian vector product 
$\nabla_\theta \cL_S(\cV) \; H_{\theta}^{-1}$ in~\equref{eq:influence}.
}
\label{tab:hessian}
\end{table}%
Computing~\equref{eq:influence}, restated here,
$$
\frac{\partial \cL_S(\cV, \theta^*(\Lambda))}{\partial \lambda_u} = -\nabla_\theta  \cL_S(\cV, \theta^*)^\top \;
H_{\theta^*}^{-1} \; \nabla_\theta \ell_{U}(u, \theta^*),
$$
is challenging as it involves an inverse Hessian. When using a deep net, the dimension of the Hessian is potentially in the millions, which demands a lot of memory and computing resources. Prior works, summarized in~\tabref{tab:hessian}, have proposed various approximations to mitigate the computational  challenges. For example,~\citet{luketina2016scalable} propose to use an identity matrix as an approximation of the inverse Hessian, and a recent method by~\citet{lorraine2019opt} uses Neumann series to trade-off  computational resources for the quality of the approximation. Different from these approximations, our approach has lower computation time and memory usage for tuning per-example weights. For more details please refer to the ablation studies, specifically \secref{exp: analysis} in the main paper.

\section{Additional Details for Per-example Gradient Computation}
\label{sec:supp_per_exp}
In the main paper, we discussed efficient  computation of per-example gradients  and presented the details for a fully connected layer. In this section, we will provide the details for two more layers,  convolution layers and batch-norm.

{\bf Convolutional Layer.} The convolution layer can be reformulated as a fully-connected layer. Hence, theoretically, we can apply the same implementation. In practice, we found that reshaping to a fully connected layer is slow and memory intensive. Hence, we utilize the auto-vectorizing capability in Tensorflow~\cite{agarwal2019auto}.  More specifically, we slice a convoluation layer's activation into mini-batches of size 1 and call the backward function in parallel using \texttt{tf.vectorized\_map}. 

{\bf Batch-norm Layer.} Batch normalization is a special case of a fully-connected layer. The trainable parameters are the scalar weights and bias in the affine transformation. Thus, we can follow the  implementation used for a fully connected layer. 


{
\begin{algorithm}[h]
\begin{algorithmic}[1]
\REQUIRE $\alpha \in \mathbb{R}_{>0}$: step size
\REQUIRE $\beta_1 , \beta_2 \in [0,1)$:  exponential decay rates for computing running averages of gradient and its square
\REQUIRE $\epsilon$: a fixed small value
\REQUIRE $\cL(\Lambda)$: A stochastic loss function with parameters $\Lambda$.
\STATE Initialize $\Lambda, m, v \in \mathbb{R}^{|\cU|}$, $t$ and $\theta_0$ \\
\WHILE {not converged} 
\STATE $t \leftarrow t+1$
\STATE $g_t \leftarrow \nabla_\Lambda \cL_t(\Lambda_{t-1})$ (Compute gradient \wrt to the stochastic loss function)
\STATE $M \leftarrow \mathbf{1}[g_t \neq 0]$ (Obtain mask to block updates, $\mathbf{1}$ denotes the indicator function)
\STATE $m_t \leftarrow m_{t-1}+ (\beta_1-1) \cdot m_{t-1}  \odot M  + (1 - \beta_1)  \cdot g_t$
\STATE $v_t  \leftarrow v_{t-1}+ (\beta_2-1) \cdot v_{t-1} \odot M  + (1 - \beta_2)  \cdot g_t \odot g_t$
\STATE $\hat{m_t} \leftarrow m_t / (1-\beta_1^t) $
\STATE $\hat{v_t} \leftarrow v_t / (1-\beta_2^t) $
\STATE $\Lambda_t \leftarrow \Lambda_{t-1} - \alpha \cdot  \hat{m_t} \odot M / ( \sqrt{\hat{v_t}}+ \epsilon)$
\ENDWHILE 
\end{algorithmic}
\caption{M-Adam Optimizer. We use $\odot$ to denote element-wise vector multiplication.
}
\label{alg:aadam}
\end{algorithm}
}
\section{Additional Details about Efficient Optimizer for $\Lambda$}
\label{sec:supp_eff_opt}
We illustrate the efficient implementation for updating $\Lambda$ based on the Adam optimizer in~\algref{alg:aadam}. We named this modified version M(asked)-Adam. 
Recall, we are updating $\lambda_u\in\Lambda$ only if the loss function $\cL$ depends on $u\in\cU'$, \ie, when the example is in the sampled mini-batch. Importantly, we do not want to update the running averages of the gradients with $0$ for all examples which are \textit{not} in the mini-batch. To do so, we introduce a mask $M \triangleq \mathbf{1}[\nabla_\Lambda \cL(\Lambda) \neq 0]$ which indicates whether the gradient  \wrt  a particular $\lambda_u$ is 0. We use $\mathbf{1}[\cdot]$ to denote the indicator function.

%

\section{Implementation Details}
\label{sec:supp_implement}
We follow the setup of UDA~\cite{xie2019uda} and FixMatch~\cite{fixmatch}. We obtain datasets and model architectures from UDA's and FixMatch's publicly available implementation\footnote{\url{https://github.com/google-research/uda}}\footnote{\url{https://github.com/google-research/fixmatch}}.

\noindent\textbf{Image Classification.}
For both UDA and FixMatch, we use the same validation set of size 1024. 
We use M-Adam with constant step size of $0.01$ as discussed in \secref{sec:supp_eff_opt} to update $\Lambda$, and SGD with momentum and a step size of $0.03$ is used to optimize $\theta$.

For UDA, we set the training batch sizes for labeled and unlabeled data to 64 and 320. The model is trained for 400k steps. The first 20k iterations are the warm-up stage where only network weights $\theta$ are optimized but not $\Lambda$. We initialize $\lambda_u, \forall u \in \cU$, to $5$ for training with 250 labeled samples and $1$ for the other settings. All experiments are performed on a single NVIDIA V100 16GB GPU. The inner step $N$ is set to 100 and the step size $\eta$ is 0.01.

Following FixMatch, the training batch sizes for labeled and unlabeled data are 64 and $448 = 64\cdot 7$. The model is trained for 1024 epochs. We initialize $\lambda_u, \forall u \in \cU$, to $1$ for all experiments. The inner step $N$ is set to 512 and step size $\eta$ is 0.01. Each experiment is performed on two NVIDIA V100 16GB GPUs.

\noindent\textbf{Text Classification.}
Following UDA~\cite{xie2019uda}, the same 20 labeled examples are used. We randomly sample another 20 to be part of the validation set as UDA did not provide a validation set. The train and validation set have equal number of examples for each category.
We use the same unlabeled data split as UDA, except we exclude the examples used in the validation set. In total, we have 69,972 unlabeled samples. We fine-tune the BERT model for 10k steps with the first 1k iterations being the warm-up phase. The training batch sizes for labeled and unlabeled data are 8 and 32. We use Adam to optimize network weights $\theta$ with learning rate $2\times 10^{-5}$.  M-Adam is used to optimize $\Lambda$ with constant learning rate $0.01$, and we optimize $\Lambda$ once every 5 $\theta$  optimization steps. All experiments for text classification are performed on NVIDIA V100 32GB GPUs. As mentioned in~\secref{subsec:ssl_benchmark}, UDA uses v3-32 Cloud TPU Pods which allows  to train with larger batch sizes and longer sequence lengths. In our case, the largest memory GPUs which we have access to are the V100 32GB GPUs.

\noindent\textbf{Reparamterization for Binary Classification.} 
The text classification task contains two classes and uses cross entropy  during training. The provided network architecture of UDA predicts two logits $f_{\theta_1}(x)$ and $f_{\theta_2}(x)$ one for each class given an input $x$. While this over-parametrization doesn't hurt the classification performance, it leads to unstable computation of $H_{\theta^*}^{-1}$, as $\theta_1$ and $\theta_2$ are highly correlated.


To handle this concern, we reparametrize  the final classification layer to have
parameters $\theta' \triangleq \theta_1 - \theta_2$, and we use the logits $f_{\theta'}(x)$ and $-f_{\theta'}(x)$ in the cross-entropy loss. With this implementation, we can compute a stable inverse Hessian while obtaining the same training loss of the original parametrization.

\section{Additional Discussion on Influence Functions}
\label{sec:supp_inf_func}


Eq.~\eqref{eq:influence} is derived by assuming: (a) the training objective $\cL$ is twice-differentiable and strictly convex with respect to $\theta$, and (b) $\theta^*$ has been optimized to global optimality.  While these assumptions are violated in context of deep nets,
prior works~\cite{koh2017understanding, lorraine2019opt} have demonstrate that influence functions remain accurate despite the non-convergence and non-convexity of the model. This finding is also consistent with our experimental results: SSL tasks benefit from tuning the per-example weights via influence functions.

For completeness, we provide a standard derivation of the influence function of $\theta$,  \ie,
$\frac{\partial \theta^*(\Lambda)}{\partial \lambda_j} = -H_{\theta^*}^{-1} \; \nabla_\theta \ell_{U}(j, \theta^*)$ for an unlabeled sample $j$ below.

Recall that $\theta^*$ minimize the  loss
$$\cL(\cD, \cU, \theta, \Lambda)= \cL_S(\cD, \theta) + \sum_{u\in \cU}\lambda_u \cdot \ell_U(u, \theta).$$
We assume $\cL$ is twice-differentiable and strictly convex \wrt $\theta$. Therefore, the Hessian matrix
$ H_{\theta^*} \triangleq \nabla_\theta^2 \cL(\cD, \cU, \theta^*, \Lambda) $ is positive definite and invertible.

Let's say we increase the weight $\lambda_j$ of unlabeled sample $j$ by a small value $\epsilon$ via $\lambda_j \leftarrow \lambda_j+\epsilon$ and optimize the network using the new weights to optimality. We refer to the new optimal weights as
$$
\theta^*_{\epsilon, j} = \arg\min_{\theta} \cL_S(\cD, \theta)+ \epsilon \ell_U(j, \theta) + \sum_{u\in \cU}\lambda_u \cdot \ell_U(u, \theta) =  \arg\min_{\theta} \cL(\cD, \cU, \theta, \Lambda) + \epsilon \ell_U(j, \theta).
$$

Since $\theta^*_{\epsilon, j} $ minimizes above equation, we then have the first order optimality conditions:
$$
0 = \nabla\cL(\cD, \cU, \theta^*_{\epsilon, j}, \Lambda) + \epsilon \nabla \ell_U(j, \theta^*_{\epsilon, j}) .
$$

As $\theta^*_{\epsilon, j} \rightarrow \theta^*$ when $\epsilon \rightarrow 0$, we perform a Taylor expansion of the right-hand side:
$$
0 =  [\nabla\cL(\cD, \cU, \theta^*, \Lambda) + \epsilon \nabla \ell_U(j, \theta^*)] +
    [\nabla^2\cL(\cD, \cU, \theta^*, \Lambda) + \epsilon \nabla^2 \ell_U(j, \theta^*)] \Delta_\epsilon + \mathcal{O}(\|\Delta_\epsilon\|),
$$
where the parameter change is denoted by  $\Delta_\epsilon \triangleq \theta^*_{\epsilon, j} -  \theta^*$, and $\mathcal{O}(\|\Delta_\epsilon\|)$ captures the higher order terms.

Ignoring $\mathcal{O}(\|\Delta_\epsilon\|)$ and solving for $\Delta_\epsilon$, we have:
$$
 \Delta_\epsilon \approx -[\nabla^2\cL(\cD, \cU, \theta^*, \Lambda) + \epsilon \nabla^2 \ell_U(j, \theta^*)]^{-1}  [\nabla\cL(\cD, \cU, \theta^*, \Lambda) + \epsilon \nabla \ell_U(j, \theta^*)].
$$

Recall, $\theta^*$ minimizes $\cL$. Consequently, we have $\nabla\cL(\cD, \cU, \theta^*, \Lambda)=0$. Dropping $\mathcal{O}(\epsilon^2)$ terms, we get
$$
 \Delta_\epsilon \approx -\nabla^2\cL(\cD, \cU, \theta^*, \Lambda)^{-1}  \nabla \ell_U(j, \theta^*) \epsilon = - H_{\theta^*}^{-1}  \nabla \ell_U(j, \theta^*) \epsilon.
$$

Finally, following the definition of derivatives,
$$
\frac{\partial \theta^*}{\partial \lambda_j} = \frac{\theta^*_{\epsilon, j} -  \theta^* }{\lambda_j + \epsilon - \lambda_j} \bigg|_{\epsilon \rightarrow 0}= \frac{\partial  \Delta_\epsilon }{\partial \epsilon} \approx - H_{\theta^*}^{-1}  \nabla \ell_U(j, \theta^*),  
$$
which concludes derivation of the influence function.

\end{document}